\documentclass[10pt,onecolumn,letterpaper]{article}

\usepackage{cvpr}
\usepackage{times}
\usepackage{epsfig}
\usepackage{graphicx}
\usepackage{amsmath}
\usepackage{amssymb}
\usepackage{authblk}
\usepackage{enumitem}
\usepackage{tikz}
 \usepackage{overpic}
\usepackage{nicefrac}  

\usepackage[pagebackref=true,breaklinks=true,letterpaper=true,colorlinks,bookmarks=false]{hyperref}

\cvprfinalcopy 

\newcommand {\E} {{\mathbb{E}}}
\newcommand {\R} {{\mathbb{R}}}
\newcommand {\Hc} {{\mathcal{H}}}
\newcommand {\HH} {{\mathbb{H}}}
\newcommand {\M} {{\mathcal{M}}}
\newcommand {\ba} {{\mathcal A}}
\newcommand {\taunorm} {{| \nabla \tau|_\infty}}
\newtheorem{theorem}{Theorem}[section]
\newtheorem{lemma}[theorem]{Lemma}
\newtheorem{corollary}[theorem]{Corollary}

\def\cvprPaperID{****} 

\ifcvprfinal\fi
\begin{document}

\title{Surface Networks}

%

\author[1]{Ilya Kostrikov}
\author[1]{Zhongshi Jiang}
\author[1]{Daniele Panozzo \thanks{DP was supported in part by the NSF CAREER award IIS-1652515, a gift from Adobe, and a gift from nTopology.}}
\author[1]{Denis Zorin \thanks{DZ was supported in part by the NSF awards DMS-1436591 and IIS-1320635.}}
\author[1,2]{Joan Bruna\thanks{JB was partially supported by Samsung Electronics (Improving Deep Learning using Latent Structure) and DOA W911NF-17-1-0438. Corresponding author: \href{mailto:bruna@cims.nyu.edu}{bruna@cims.nyu.edu}}}
\affil[1]{Courant Institute of Mathematical Sciences, New York University}
\affil[2]{Center for Data Science, New York University}

\maketitle

\begin{abstract}
We study data-driven representations 
for three-dimensional triangle meshes, which are one of the prevalent objects used to represent 3D geometry. 
Recent works have developed models
that exploit the intrinsic geometry of manifolds and graphs, 
namely the Graph Neural Networks (GNNs) and its spectral variants, 
which learn from the local metric tensor via the Laplacian operator. 

Despite offering excellent sample complexity and built-in invariances,
intrinsic geometry alone is invariant to isometric deformations, making it unsuitable for  many applications.
To overcome this limitation,
we propose several upgrades to GNNs 
to leverage extrinsic differential geometry properties 
 of three-dimensional surfaces, increasing its modeling power. 
 In particular, we propose to exploit the Dirac operator, whose spectrum detects principal curvature directions --- this is in stark contrast with the classical Laplace 
 operator, which directly measures mean curvature. We coin the 
 resulting models \emph{Surface Networks (SN)}.

We prove that these models define shape representations that are stable to deformation and to discretization, and we demonstrate the efficiency and versatility of SNs on 
 two
 challenging tasks: temporal prediction of mesh deformations
 under non-linear dynamics and generative models using 
 a variational autoencoder framework with encoders/decoders
 given by SNs.
 
\end{abstract}

 \section{Introduction}


3D geometry analysis, manipulation and synthesis plays an important role in a variety of applications from engineering to computer animation to medical imaging. 
Despite the vast amount of high-quality 3D geometric data available, data-driven approaches to problems involving complex geometry have yet to become mainstream, in part due to the lack of data representation regularity which is required for traditional convolutional neural network approaches. While  in computer vision problems inputs are typically sampled on regular 
two or three-dimensional grids, surface geometry is represented in a more complex form and, in general, cannot be  converted to an image-like format by parametrizing the shape using a single planar chart.  
Most commonly an irregular triangle mesh is used to represent shapes, capturing its main topological and geometrical properties.

Similarly to the regular grid case (used for images or videos), we are interested in data-driven 
representations that strike the right balance between expressive power and sample complexity. 
In the case of CNNs, this is achieved by exploiting the inductive bias that most computer 
vision tasks are locally stable to deformations, leading to localized, multiscale, stationary features. 
In the case of surfaces, we face a fundamental modeling choice between \emph{extrinsic} versus \emph{intrinsic} representations. 
Extrinsic representations rely on the specific embedding of surfaces within a three-dimensional ambient space, 
whereas intrinsic representations only capture geometric properties specific to the surface, irrespective of 
its parametrization. Whereas the former offer arbitrary representation power, they are unable to easily exploit inductive priors such as stability to local deformations and invariance to global transformations. 

A particularly simple and popular extrinsic method \cite{qi2016pointnet, qi2017pointnet++} represents shapes as point clouds in $\R^3$ of variable size, and leverages recent deep learning models that operate on input sets \cite{vinyals2015order, sukhbaatar2016learning}. Despite its advantages in terms of ease of data acquisition (they no longer require a mesh triangulation) and good empirical performance on shape classification and segmentation tasks, one may wonder whether this simplification comes at a loss of precision as one considers more challenging prediction tasks.


In this paper, we develop an alternative pipeline that applies neural networks directly on triangle meshes, building 
on \emph{geometric deep learning}. These models provide 
data-driven intrinsic graph and manifold representations with inductive biases analogous to CNNs on natural images.
Models based on Graph Neural Networks \cite{scarselli2009graph} and their spectral 
variants \cite{bruna2013spectral, defferrard2016convolutional, kipf2016semi} have been successfully applied to geometry processing tasks such 
as shape correspondence \cite{monti2016geometric}. In their basic form, these models learn a deep representation over the discretized surface by combining a latent representation at a given node with a local linear combination 
of its neighbors' latent representations, and a point-wise nonlinearity.
 Different models vary in their choice of linear operator and point-wise nonlinearity, 
 which notably includes the graph Laplacian, leading to spectral interpretations of those models.

Our contributions are three-fold. First, we extend the model to support extrinsic features.
 More specifically, we exploit the fact that surfaces in 
$\R^3$ admit a first-order differential operator, the \emph{Dirac} operator, that
is stable to discretization, provides a direct generalization of Laplacian-based 
propagation models, and is able to detect principal curvature directions \cite{crane2011spin, Liu:Dirac:2017}. 
Next, we prove that the models resulting from either Laplace or Dirac operators are stable 
to deformations and to discretization, two major sources of variability in practical applications. 
Last, we introduce a generative model for surfaces based on the variational autoencoder framework \cite{kingma2013auto, rezende2015variational}, that is able to exploit non-Euclidean geometric regularity. 

By combining the Dirac operator with input coordinates, we obtain a fully differentiable,
end-to-end feature representation that we apply to several challenging tasks. 
The resulting Surface Networks -- using either the Dirac or the Laplacian, inherit 
the stability and invariance properties of these operators, thus providing data-driven representations with certified stability to deformations. 
We demonstrate the model efficiency on a temporal prediction task of complex dynamics, based on a physical simulation of elastic shells, which confirms that whenever geometric information (in the form of a mesh) is available, it can be leveraged to significantly outperform point-cloud based models. 


Our main contributions are summarized as follows:
\begin{itemize}
\item We demonstrate that Surface Networks provide accurate temporal prediction of surfaces under complex non-linear dynamics, motivating the use of geometric shape information.
\item We prove that Surface Networks define shape representations that are stable to deformation and to discretization.
\item We introduce a generative model for 3D surfaces based on the variational autoencoder. 
\end{itemize}

A reference implementation of our algorithm is available at \url{https://github.com/jiangzhongshi/SurfaceNetworks}.



%

 \section{Related Work}


Learning end-to-end representations on irregular and non-Euclidean domains 
is an active and ongoing area of research. \cite{scarselli2009graph} introduced graph 
neural networks as recursive neural networks on graphs, whose stationary 
distributions could be trained by backpropagation. Subsequent works \cite{li2015gated, sukhbaatar2016learning} have relaxed the model by untying the recurrent layer weights and proposed several nonlinear updates through gating mechanisms. 
Graph neural networks are in fact natural generalizations of convolutional networks to non-Euclidean graphs. \cite{bruna2013spectral, henaff2015deep} proposed to learn smooth spectral multipliers of the graph Laplacian, albeit with high computational cost, and \cite{defferrard2016convolutional, kipf2016semi} resolved the computational bottleneck by learning polynomials of the graph Laplacian, thus avoiding the computation of eigenvectors and completing the connection with GNNs. We refer the reader to \cite{bronstein2016geometric} for an exhaustive literature review on the topic. 
GNNs are finding application in many different domains. \cite{battaglia2016interaction, torralba16} develop graph interaction networks that learn pairwise particle interactions and apply them to discrete particle physical dynamics. \cite{duvenaud2015convolutional, kearnes2016molecular} study molecular fingerprints using variants of the GNN architecture, and \cite{gilmer2017neural} further develop the model by combining it with set representations \cite{vinyals2015order}, showing state-of-the-art results on molecular prediction. The resulting models, so-called Message-Passing Neural Networks, also learn the diffusion operator, which can be seen as generalizations of the Dirac model on general graphs. 

In the context of computer graphics, \cite{masci2015geodesic} developed the first CNN model on meshed surfaces using intrinsic patch representations, and further generalized in \cite{boscaini2016learning} and \cite{monti2016geometric}. This last work allows for flexible representations via the so-called pseudo-coordinates and obtains state-of-the-art results on 3D shape correspondence, although it does not easily encode first-order differential information.
These intrinsic models contrast with Euclidean models such as \cite{wu20153d, wei2016dense}, that have higher sample complexity, since they need to learn the underlying invariance of the surface embedding.
Point-cloud based models are increasingly popular to model 3d objects due to their simplicity and versatility. \cite{qi2016pointnet,qi2017pointnet++} use set-invariant representations from \cite{vinyals2015order,sukhbaatar2016learning} to solve shape segmentation and classification tasks. 
More recently, \cite{toralcnn_siggraph17} proposes to learn surface convolutional network from a canonical representation of planar flat-torus, with excellent performance on shape segmentation and classification, although such canonical representations may introduce exponential scale changes that  can introduce instabilities.  Finally, \cite{fan2016point} proposes a point-cloud generative model for 3D shapes, that incorporates invariance to point permutations, but does not encode geometrical information as our shape generative model. Learning variational deformations is an important problem for graphics applications, since it enables negligible and fixed per-frame cost \cite{Poranne:2015}, but it is currently limited to 2D deformations using point handles. In constrast, our method easily generalizes to 3D and learns dynamic behaviours.








 \section{Surface Networks}
\label{snnsec}

This section presents our surface neural network model
and its basic properties. We start by introducing the 
problem setup and notations using the Laplacian formalism (Section \ref{sec:laplacian}), 
and then introduce our model based on the Dirac operator (Section \ref{sec:dirac}).

\subsection{Laplacian Surface Networks}
\label{sec:laplacian}
Our first goal is to define a trainable representation of 
discrete surfaces. Let $\M=\{V,E,F\}$ be a triangular 
mesh, where $V=(v_i \in \R^3)_{i \leq N}$ contains the node coordinates, 
$E = (e_{i,j} )$ corresponds to edges, and $F$ is the set of triangular faces.
We denote as $\Delta$ the discrete Laplace-Beltrami operator (we use the popular  cotangent weights formulation, see \cite{bronstein2016geometric} for details).

This operator can be interpreted as a local, linear high-pass filter in $\M$ 
that acts on signals $x \in \R^{d \times |V|}$ defined on the vertices
as a simple matrix multiplication $\tilde{x} = \Delta x$.
By combining $\Delta$ with an \emph{all-pass} filter and learning generic 
linear combinations followed by a point-wise nonlinearity, we obtain 
a simple generalization of localized convolutional operators in $\M$ 
that update a feature map from layer $k$ to layer $k+1$ using trainable parameters $A_k$ and $B_k$:
\begin{equation}
\label{laplacenet}
x^{k+1} = \rho \left( A_k \Delta x^k + B_k x^k \right)~,~A_k,B_k \in \R^{d_{k+1} \times d_k} ~.
\end{equation}

By observing that the Laplacian itself can be written in terms of the graph weight similarity 
by diagonal renormalization, this model is a specific instance of the graph neural network \cite{scarselli2009graph, bronstein2016geometric, kipf2016semi} 
and a generalization of the spectrum-free Laplacian networks from \cite{defferrard2016convolutional}. 
As shown in these previous works, convolutional-like layers (\ref{laplacenet}) can be 
combined with graph coarsening or pooling layers. 

In contrast to general graphs, meshes contain a low-dimensional Euclidean embedding that 
contains potentially useful information in many practical tasks, despite being extrinsic and thus 
not invariant to the global position of the surface. A simple strategy to strike a good balance between 
expressivity and invariance is to include the node canonical coordinates as input channels to the network: $x^{1}:=V \in \R^{|V| \times 3}$. 
The mean curvature can be computed by applying the Laplace operator to the coordinates of the vertices: 
\begin{equation}
\label{laplacenorm}
\Delta x^{1} = -2 H \bf{n}~,
\end{equation}
where $H$ is the mean curvature function and ${\bf n}(u)$ is the normal vector of the surface at point $u$.
As a result, the Laplacian neural model (\ref{laplacenet}) has access to mean curvature and normal information.
Feeding Euclidean embedding coordinates into graph neural network models is related to the use of generalized coordinates from \cite{monti2016geometric}.
By cascading $K$ layers of the form (\ref{laplacenet}) we obtain a representation $\Phi_{\Delta}(\M)$ 
that contains generic features at each node location. When the number of layers $K$ is of the order of 
$\text{diam}(\M)$, the diameter of the graph determined by $\M$, then the network is able to propagate and aggregate
information across the whole surface.

Equation (\ref{laplacenorm}) illustrates that a Laplacian layer is only able to extract isotropic high-frequency information, 
corresponding to the mean variations across all directions. Although in general graphs there is no well-defined procedure 
to recover anisotropic local variations, in the case of surfaces some authors (\cite{boscaini2016learning, andreux2014anisotropic,monti2016geometric}) have 
considered anisotropic extensions. We describe next a particularly simple procedure to increase 
the expressive power of the network using a related operator from quantum mechanics: the Dirac operator, that has been previously used successfully in the context of surface deformation \cite{crane2011spin} and shape analysis \cite{Liu:Dirac:2017}.
 
\subsection{Dirac Surface Networks}

\label{sec:dirac}

The Laplace-Beltrami operator $\Delta$ is a second-order differential operator, 
constructed as $\Delta = -\text{div} \nabla$ by combining the gradient (a first-order differential 
operator) with its adjoint, the divergence operator. In an Euclidean space, one has access to 
these first-order differential operators separately, enabling oriented high-pass filters. 

For convenience, we embed $\R^3$ to the imaginary quaternion space $\text{Im}(\HH)$ (see Appendix A in the Suppl. Material for details). 
The Dirac operator is then defined as a matrix $D  \in \HH^{|F| \times |V|}$ that maps (quaternion) signals on the nodes to signals on the faces. 
In coordinates, 
$$D_{f,j} = \frac{-1}{2 | \ba_f | }e_j~,~f \in F, j \in V~,$$
where $e_j$ is the opposing edge vector of node $j$ in the face $f$, and $\ba_f$ is 
the area
(see Appendix A)
using counter-clockwise orientations on all faces. 


To apply the Dirac operator defined in quaternions to signals in vertices and faces defined in real numbers, we write the feature vectors as quaternions by splitting them into chunks of 4 real numbers representing the real and imaginary parts of a quaternion; see Appendix A. Thus, we always work with feature vectors with dimensionalities that are multiples of $4$. The Dirac operator provides first-order differential information 
and is sensitive to local orientations. 
Moreover, one can verify \cite{crane2011spin} that 
$$ \mbox{Re } D^* D = \Delta~,$$
where $D^*$ is the adjoint operator of $D$ in the quaternion space (see Appendix A).
The adjoint matrix can be computed as $D^* = M^{-1}_V D^H M_F$ where $D^H$ is a conjugate transpose of $D$ and $M_V$, $M_F$ are diagonal mass matrices with one third of areas of triangles incident to a vertex and face areas respectively.

The Dirac operator can be used to define a new neural surface representation 
that alternates layers with signals defined over nodes with layers defined over faces. 
Given a $d$-dimensional feature representation over the nodes $x^k \in \R^{d \times |V|}$, and the faces of the mesh, $y^k \in \R^{d \times |F|}$,
we define a $d'$-dimensional mapping to a face representation as 
\begin{equation}
\label{eq:dir1}
y^{k + 1}=  \rho \left(C_k D x^k + E_k y^k \right), C_k, E_k \in \R^{d_{k+1} \times d_k},
\end{equation}
where $C_k, E_k$ are trainable parameters.
Similarly, we define the adjoint layer that maps back to a $\tilde{d}$-dimensional signal over nodes as
\begin{equation}
\label{eq:dir2}
x^{k+1} = \rho \left(A_k D^*y^{k+1} + B_k x^k \right)~, A_k, B_k \in \R^{d_{k+1} \times d_k},
\end{equation}
where $A_k, B_k$ are trainable parameters.
A surface neural network layer is thus determined by parameters $\{A, B, C, E\}$ using equations (\ref{eq:dir1}) and (\ref{eq:dir2}) 
to define $x^{k+1} \in \R^{d_{k+1} \times |V|}$.  We denote by $\Phi_D(\M)$ the mesh representation resulting 
from applying $K$ such layers (that we assume fixed for the purpose of exposition).



The Dirac-based surface network is related to 
edge feature transforms proposed on general graphs in \cite{gilmer2017neural}, 
although these edge measurements cannot be associated with derivatives 
due to lack of proper orientation. In general graphs, there is no notion of square root of $\Delta$ that recovers oriented first-order derivatives. 


\section{Stability of Surface Networks}
\label{stabsection}

Here we describe how Surface Networks are geometrically stable, 
because surface deformations become additive noise under the model. 
Given a continuous surface $S \subset \R^3$ or a discrete mesh $\M$, and a smooth deformation field $\tau: \R^3 \to \R^3$, 
we are particularly interested in two forms of stability: 
\begin{itemize} 
\item Given a discrete mesh $\M$ and a certain non-rigid deformation $\tau$ 
acting on $\M$, we want to certify that $ \| \Phi(\M) - \Phi(\tau(\M)) \|$ is 
small if $\| \nabla \tau ( \nabla \tau)^* - {\bf I} \| $ is small, i.e when the deformation is nearly rigid; see Theorem \ref{stabtheo}.
 \item Given two discretizations $\M_1$ and $\M_2$ of the same underlying surface $S$, 
we would like to control $\| \Phi( \M_1) - \Phi(\M_2) \| $ in terms of the resolution of the meshes; see Theorem \ref{stabtheo2}.
\end{itemize}
These stability properties are important in applications, since most tasks we are interested in 
are stable to deformation and to discretization. We shall see that the first property is a simple 
consequence of the fact that the mesh Laplacian and Dirac operators are themselves
stable to deformations. The second property will require us to specify under 
which conditions the discrete mesh Laplacian $\Delta_\M$ converges to the 
Laplace-Beltrami operator $\Delta_S$ on $S$. Unless it is clear from the 
context, in the following $\Delta$ will denote the discrete Laplacian.
\begin{theorem}
\label{stabtheo}
Let $\M$ be a $N$-node mesh and $x,\,x' \in \R^{|V| \times d}$ be 
 input signals defined on the nodes. Assume the nonlinearity $\rho(\,\cdot \,)$ is 
 non-expansive ($ | \rho(z) - \rho(z') | \leq | z - z'|$). Then
\begin{enumerate}[label=(\alph*)]
\item $\| \Phi_\Delta(\M; x) - \Phi_\Delta(\M; x') \| \leq \alpha_\Delta \| x - x' \|~,$ 
where $\alpha_\Delta$ depends only on the trained weights and the mesh.
\item $\| \Phi_D(\M; x) - \Phi_D(\M; x') \| \leq \alpha_D \| x - x' \|~,$
where $\alpha_D$ depends only on the trained weights and the mesh.
\item Let $\taunorm := \sup_u \| \nabla \tau(u) (\nabla \tau(u))^* - {\bf 1} \|$, where $\nabla \tau(u)$ is the Jacobian
matrix of $u \mapsto \tau(u)$. Then $\| \Phi_\Delta(\M; x) - \Phi_\Delta( \tau(\M); x) \| \leq \beta_\Delta \taunorm \|x \|~,$
where $\beta_\Delta$ is independent of $\tau$ and $x$.
\item Denote by $\widetilde{|\nabla \tau |}_\infty := \sup_u \| \nabla \tau(u) - {\bf 1} \|$. Then 
$\| \Phi_D(\M; x) - \Phi_D( \tau(\M); x) \| \leq \beta_D \widetilde{|\nabla \tau |}_\infty  \|x \|~,$
where $\beta_D$ is independent of $\tau$ and $x$.
\end{enumerate}
\end{theorem}
Properties (a) and (b) are not specific to surface representations, and 
are a simple consequence of the non-expansive property of our chosen 
nonlinearities. The constant $\alpha$ is controlled by the product 
of $\ell_2$ norms of the network weights at each layer and the 
norm of the discrete Laplacian operator. 
Properties (c) and (d) are based on the fact that the Laplacian and Dirac operators 
are themselves stable to deformations, a property that 
depends on two key aspects: first, the Laplacian/Dirac is localized 
in space, and next, that it is a high-pass filter and therefore only depends 
on relative changes in position.

One caveat of Theorem \ref{stabtheo} is that the constants 
appearing in the bounds depend upon a bandwidth parameter given by 
the reciprocal of triangle areas, which increases
as the size of the mesh increases. This corresponds to the 
fact that the spectral radius of $\Delta_\M$ diverges as the mesh 
size $N$ increases.

In order to overcome this problematic asymptotic behavior, it is necessary to exploit the 
smoothness of the signals incoming to the surface network. This can be measured 
with Sobolev norms defined using the spectrum of the Laplacian operator.
Given a mesh $\M$ of $N$ nodes approximating an underlying surface $S$, and its associated cotangent Laplacian $\Delta_\M$, consider the spectral decomposition of $\Delta_\M$ (a symmetric, positive definite operator):
$$\Delta_\M = \sum_{k \leq N} \lambda_k e_k e_k^T~,~e_k \in \R^N~,~0 \leq \lambda_1 \leq \lambda_2 \dots \leq \lambda_N~. $$
Under normal uniform convergence \footnote{which controls how the normals
 of the mesh align with the surface normals; see \cite{wardetzky2008convergence}.} \cite{wardetzky2008convergence}, the spectrum of $\Delta_\M$ 
converges to the spectrum of the Laplace-Beltrami operator $\Delta_S$ of $S$.
If $S$ is bounded, it is known from the Weyl law \cite{weyl1911asymptotische} that there exists $\gamma>0$ such that
$k^{-\gamma(S)} \lesssim \lambda_k^{-1}$, so the eigenvalues $\lambda_k$ do not grow too fast.
The smoothness of a signal $x \in \R^{|V| \times d}$ defined in $\M$ is captured 
by how fast its spectral decomposition $\hat{x}(k) = e_k^T x \in \R^d$ decays \cite{spielman2007spectral}.
We define $\|x \|_{\Hc}^2:= \sum_k \lambda(k)^2 \|\hat{x}(k) \|^2$ is Sobolev norm,
and $\beta(x,S) >1$ as the largest rate 
such that its spectral decomposition coefficients  satisfy
\begin{equation}
\label{betadef}
\|\hat{x}(k) \| \lesssim k^{-\beta} ~,~(k \to \infty)~.
\end{equation}
If $x \in \R^{|V| \times d}$ is the input to the Laplace Surface Network of $R$ layers,
we denote by $(\beta_0, \beta_1, \dots, \beta_{R-1})$ the smoothness rates of the feature 
maps $x^{(r)} $ defined at each layer $r \leq R$.  


\begin{theorem}
\label{stabtheo2}
Consider a surface $S$ and a finite-mesh approximation $\M_N$ of $N$ points, 
and $\Phi_\Delta$ a Laplace Surface Network with parameters $\{(A_r, B_r)\}_{r \leq R}$. 
Denote by $d(S, \M_N)$ the uniform normal distance, and
let $x_1,x_2$ be piece-wise polyhedral approximations of $\bar{x}(t)$, $t \in S$ in $\M_N$,
with $\| \bar{x} \|_{\Hc(S)} < \infty$. Assume $ \| \bar{x}^{(r)} \|_{\Hc(S)} < \infty$ 
for $r \leq R$. 
\begin{enumerate}[label=(\alph*)]
\item If $x_1,x_2$ are two functions such that the $R$ feature maps $x_l^{(r)}$ have rates $(\beta_0, \beta_1, \dots, \beta_{R-1})$, then
\begin{equation}
\label{pony1}
\| \Phi_\Delta(x_1;\M_N) - \Phi_\Delta(x_2;\M_N) \|^2 \leq C(\beta)  \| x_1 - x_2\|^{h(\beta)} ~, 
\end{equation}
with $h(\beta) = {\prod_{r=1}^R \frac{\beta_r-1}{\beta_r-1/2}}$, 
and where $C(\beta)$ does not depend upon $N$. 
\item If $\tau$ is a smooth deformation field, then 
$ \| \Phi_\Delta(x; \M_N) - \Phi_\Delta(x; \tau(\M_N)) \| \leq C \taunorm^{h(\beta)}~, $
where $C$ does not depend upon $N$. 
\item Let $\M$ and $\M'$ be $N$-point discretizations of $S$, 
If $\max(d(\M, S), d(\M',S) ) \leq \epsilon$, then
$\| \Phi_\Delta(\M;x) - \Phi_\Delta(\M', x') \| \leq C \epsilon^{h(\beta)} ~,$
where $C$ is independent of $N$.  
\end{enumerate}
\end{theorem}

This result ensures that if we use as generator 
of the SN an operator that is consistent as the mesh resolution 
increases, the resulting surface representation is also consistent. 
Although our present result only concerns the Laplacian, the Dirac 
operator also has a well-defined continuous counterpart \cite{crane2011spin}
that generalizes the gradient operator in quaternion space.
Also, our current bounds depend explicitly upon the smoothness of feature maps
across different layers, which may be controlled in terms of the 
original signal if one considers nonlinearities that demodulate 
the signal, such as $\rho(x) = |x|$ or $\rho(x) = \text{ReLU}(x)$. These extensions are left for future work.
Finally, a specific setup that we use in experiments is to 
use as input signal the canonical coordinates of the mesh $\M$.
In that case, an immediate application of the previous theorem yields
\begin{corollary}
\label{corocombine}
Denote $\Phi(\M) := \Phi_{\M}(V)$, where $V$ are the node coordinates of $\M$. 
Then, if $A_1 =0 $, 
\begin{equation}
\| \Phi(\M) - \Phi(\tau(\M)) \| \leq \kappa \max(\taunorm, \| \nabla^2 \tau \|)^{h(\beta)}~.
\end{equation}
\end{corollary}




 \section{Generative Surface Models}
\label{genmodelsec}

\begin{figure}
\centering
\includegraphics[width=0.5\textwidth]{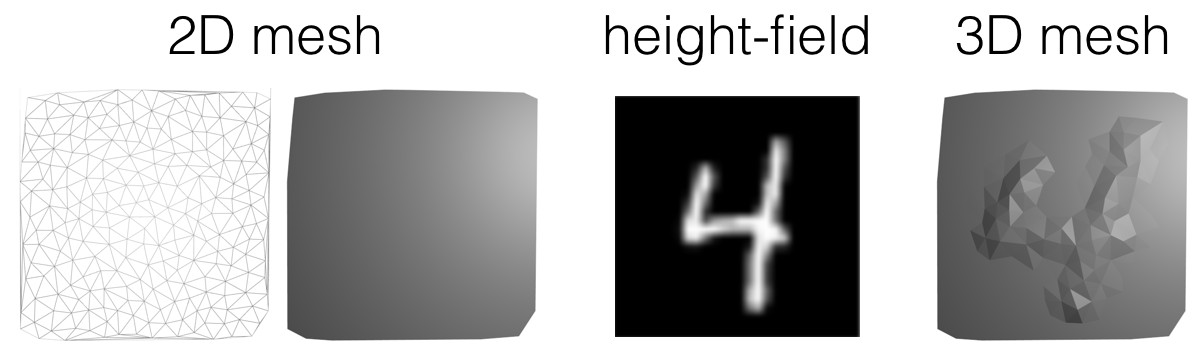}
\caption{Height-Field Representation of surfaces. A 3D mesh $\M \subset \R^3$  (right) is expressed in terms of a ``sampling" 2D irregular mesh $\tilde{\M} \subset \R^2$ (left) and a depth scalar field $f:\tilde{\M} \to \R$ over $\tilde{\M}$ (center). }  
\label{fig:genfigure0}
\end{figure}

State-of-the-art generative models for images, 
such as generative adversarial networks \cite{radford2015unsupervised}, pixel autoregressive 
networks \cite{oord2016pixel}, or variational autoencoders \cite{kingma2013auto}, exploit the locality 
and stationarity of natural images in their probabilistic models,
in the sense that the model satisfies $p_\theta(x) \approx p_\theta( x_\tau)$
by construction, where $x_\tau$ is a small deformation of a given input $x$. 
This property is obtained via encoders and decoders 
with a deep convolutional structure.  
We intend to exploit similar geometric stability priors 
with SNs, owing to their stability properties described in Section \ref{stabsection}.
A mesh generative model contains two distinct sources of randomness: 
on the one hand, the randomness associated with the underlying continuous surface, 
which corresponds to shape variability; on the other hand, 
the randomness of the discretization of the surface. Whereas 
the former contains the essential semantic information, the latter is not 
informative, and to some extent independent of the shape identity. 
We focus initially 
on meshes that can be represented as a depth map over an (irregular) 
2D mesh, referred as \emph{height-field} meshes in the literature.
That is, a mesh $\M=(V, E, F)$ is expressed as 
$(\tilde{\M}, f(\tilde{\M}))$, where $\tilde{\M}=(\tilde{V}, \tilde{E}, \tilde{F})$ is now a 2D mesh and $f~:~\tilde{V} \to \R$ 
is a \emph{depth}-map encoding the original node locations $V$, as shown in Figure \ref{fig:genfigure0}.

In this work, we consider the variational autoencoder framework \cite{kingma2013auto, rezende2015variational}. 
It considers a mixture model of the form $p(\M) = \int p_\theta(\M ~|~h) p_0(h) dh~$,
where $h \in \R^|S|$ is a vector of latent variables. 
We train this model by optimizing the variational lower bound of the data log-likelihood:
{\small 
\begin{equation}
\min_{\theta, \psi} \frac{1}{L} \sum_{l \leq L} - \E_{ h \sim q_\psi(h ~|~\M_l)} \log p_\theta( \M_l ~|~h) + D_{KL}( q_\psi( h~|~\M_l) ~||~p_0(h) )~.
\end{equation}
}
We thus need to specify a conditional generative model $p_\theta(\M ~|~h)$, 
a prior distribution $p_0(h)$ and a variational approximation to the posterior
$q_\psi( h ~|~ \M)$, where $\theta$ and $\psi$ denote respectively generative 
and variational trainable parameters.
Based on the height-field representation, we choose for simplicity a separable model of the form
$p_\theta(\M ~|~h) =  p_\theta( f~|~h, \tilde{\M}) \cdot p(\tilde{\M}) ~,$
where $\tilde{\M} \sim p(\tilde{\M})$ is a homogeneous Poisson point process, 
and $f \sim p_\theta( f~|~h, \tilde{\M})$ is a normal distribution with 
mean and isotropic covariance parameters given by a SN:
$$p_\theta( f~|~h, \tilde{\M}) = {\cal N}( \mu(h, \tilde{\M}), \sigma^2(h, \tilde{\M}) {\bf 1})~,$$
with  $[ \mu(h, \tilde{\M}), \sigma^2(h, \tilde{\M})] = \Phi_D(\tilde{M}; h)~.$
The generation step thus proceeds as follows.
We first sample a 2D mesh $\tilde{\M}$ independent of the latent variable $h$, 
and then sample a depth field over $\tilde{\M}$ conditioned on $h$ from the output
of a decoder network $\Phi_D(\tilde{M}; h)$.
Finally, the variational family $q_\psi$ is also a Normal distribution whose parameters 
are obtained from an encoder Surface Neural Network whose last layer is a global pooling that removes 
the spatial localization:
$q_\psi( h~|~\M) = {\cal N}(\bar{\mu}, \bar{\sigma}^2 {\bf 1})~,~\text{ with }~ [\bar{\mu}, \bar{\sigma}] = \bar{\Phi}_D(\M)~.$



 \section{Experiments}

\begin{figure}
\vspace*{-5mm}
\centering
\includegraphics[width=0.49\textwidth]{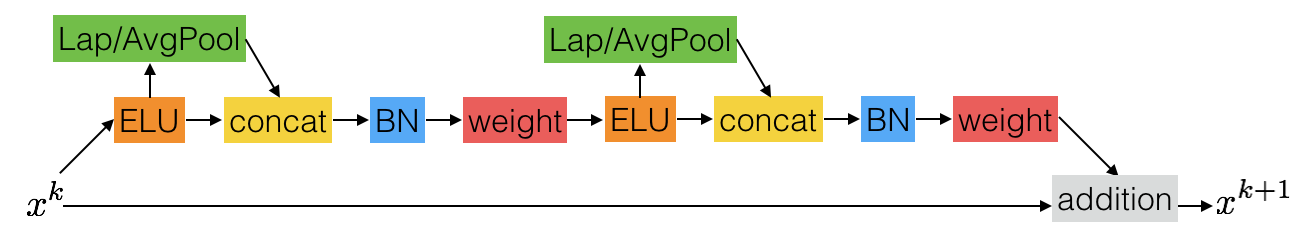}
\includegraphics[width=0.49\textwidth]{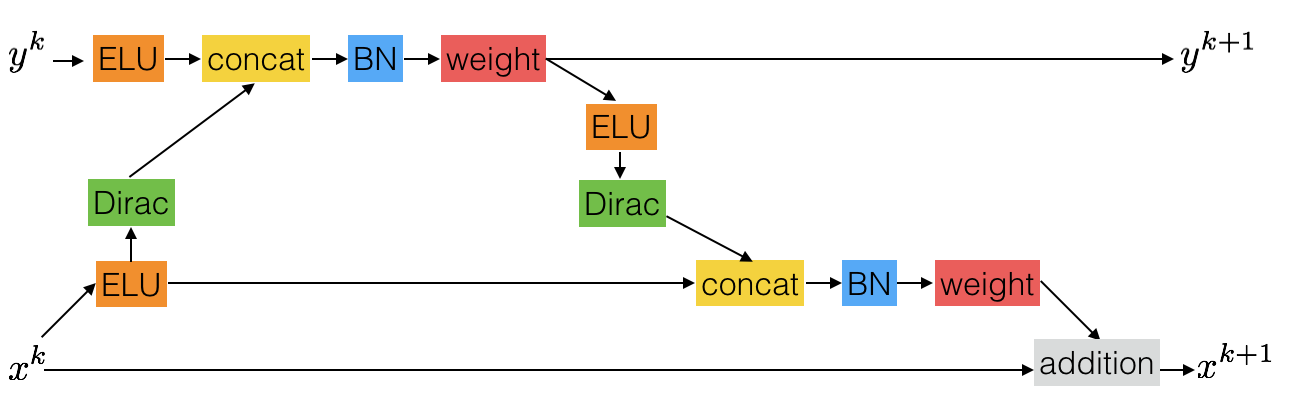}
\caption{A single ResNet-v2 block used for Laplace, Average Pooling (top) and Dirac models (bottom). The green boxes correspond to the linear operators replacing convolutions in regular domains. We consider Exponential Linear Units (ELU) activations (orange), Batch Normalization (blue) and `$1 \times 1$' convolutions (red) containing the trainable parameters; see Eqs (\ref{laplacenet}, \ref{eq:dir1} and \ref{eq:dir2}). We slightly abuse language and denote by $x^{k+1}$ the output of this 2-layer block.}  \label{fig:resnet}
\end{figure}

For experimental evaluation, we compare models built using ResNet-v2 blocks \cite{he2016identity}, where convolutions are replaced with the appropriate operators (see Fig. \ref{fig:resnet}): \textit{(i)} a point cloud based model from \cite{sukhbaatar2016learning} that aggregates global information by averaging features in the intermediate layers and distributing them to all nodes; \textit{(ii)} a Laplacian Surface network with input canonical coordinates; \textit{(iii)} a Dirac Surface Network model. We report experiments on generative models using an unstructured variant of MNIST digits (Section \ref{genmodelexp}), and on temporal prediction under non-rigid deformation models (Section \ref{spatiotemporalsect}). 


\subsection{MeshMNIST}
\label{genmodelexp}

For this task, we construct a MeshMNIST database with only \emph{height-field} meshes (Sec. \ref{genmodelsec}). First, we  sample points on a 2D plane $([0, 27] \times [0, 27])$ with Poisson disk sampling with $r=1.0$, which roughly generates $500$ points, and apply a Delaunay triangulation to these points. We then overlay the triangulation with the original MNIST images and assign to each point a $z$ coordinate bilinearly interpolating the grey-scale value. Thus, the procedure allows us to define a sampling process over 3D height-field meshes.

\begin{figure}[!ht]
    \vspace*{-2mm}
\centering
 \includegraphics[width=0.09\textwidth]{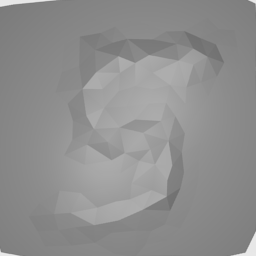} 
 \includegraphics[width=0.09\textwidth]{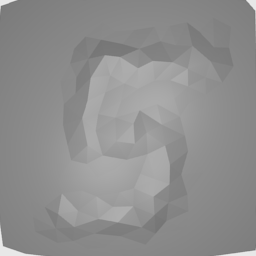} 
 \includegraphics[width=0.09\textwidth]{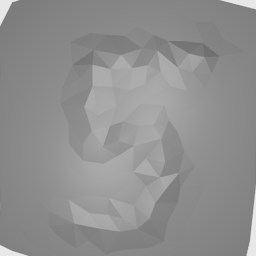} 
 \includegraphics[width=0.09\textwidth]{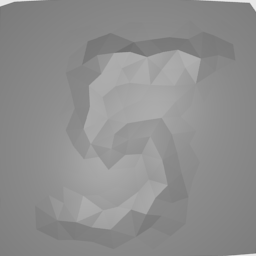} 
 \includegraphics[width=0.09\textwidth]{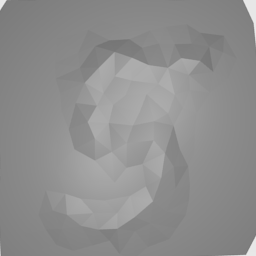} 
 \includegraphics[width=0.09\textwidth]{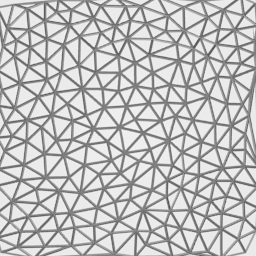} 
 \includegraphics[width=0.09\textwidth]{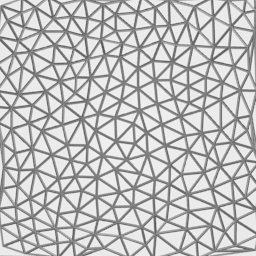} 
 \includegraphics[width=0.09\textwidth]{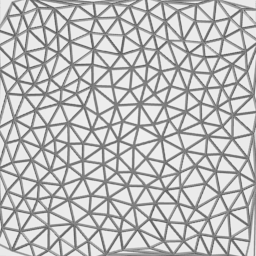} 
 \includegraphics[width=0.09\textwidth]{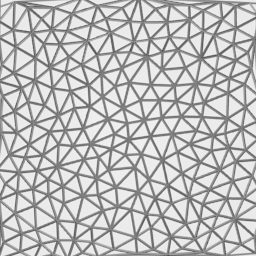} 
 \includegraphics[width=0.09\textwidth]{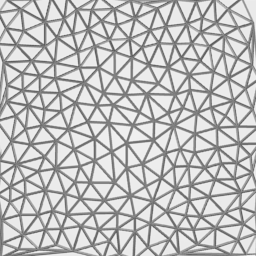} \\
\caption{Samples generated for the same latent variable and different triangulations. The learned representation is independent of discretization/triangulation (Poisson disk sampling with p=1.5).
} 
\label{fig:mnist_tri}
\end{figure}

\begin{figure}[!ht]
    \vspace*{-2mm}
\centering
 \includegraphics[width=0.09\textwidth]{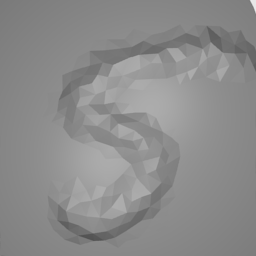} 
 \includegraphics[width=0.09\textwidth]{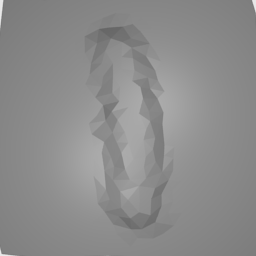} 
 \includegraphics[width=0.09\textwidth]{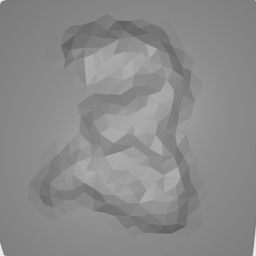} 
 \includegraphics[width=0.09\textwidth]{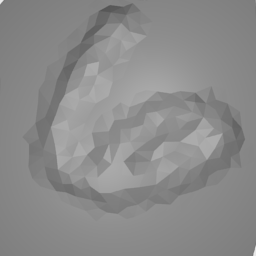} 
 \includegraphics[width=0.09\textwidth]{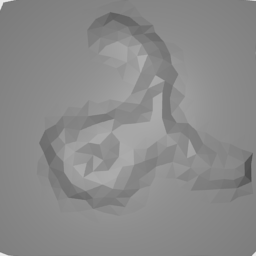} 
 \includegraphics[width=0.09\textwidth]{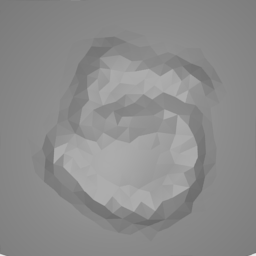} 
 \includegraphics[width=0.09\textwidth]{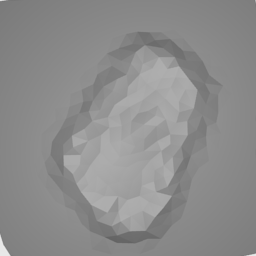} 
 \includegraphics[width=0.09\textwidth]{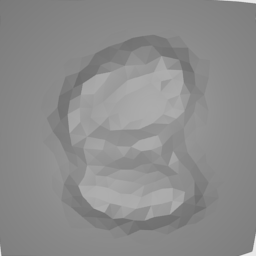} 
 \includegraphics[width=0.09\textwidth]{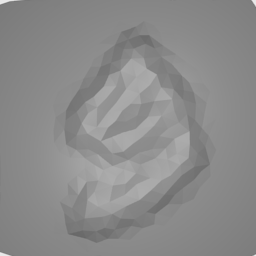} 
 \includegraphics[width=0.09\textwidth]{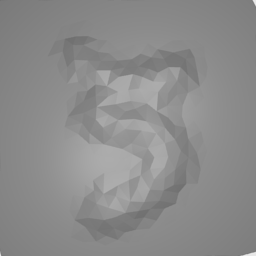} 
\caption{Meshes from the dataset (first five). And meshes generated by our model (last five).}  
\label{fig:mnist_real_fake}
\end{figure}

We used VAE models with decoders and encoders built using 10 ResNet-v2 blocks with 128 features.  The encoder converts a mesh into a latent vector by averaging output of the last ResNet-v2 block and applying linear transformations to obtain mean and variance, while the decoder takes a latent vector and a 2D mesh as input (corresponding to a specific 3D mesh) and predicts offsets for the corresponding locations. We keep variance of the decoder as a trainable parameter that does not depend on input data.
We trained the model for $75$ epochs using Adam optimizer \cite{kingma2015adam} with learning rate $10^{-3}$, weight decay $10^{-5}$ and batch size $32$. Figures \ref{fig:mnist_tri},\ref{fig:mnist_real_fake} illustrate samples from the model. The geometric encoder is able to leverage the local translation invariance of the data despite the irregular sampling, whereas the geometric decoder automatically adapts to the specific sampled grid, as opposed to set-based generative models. 

\subsection{Spatio-Temporal Predictions}
\label{spatiotemporalsect}

\begin{table*}
    \centering
    \begin{tabular}{ l | c  c  ||r }
    \hline
      Model & Receptive field & Number of parameters & Smooth L1-loss (mean per sequence (std)) \\    \hline
      MLP & 1 & 519672 & 64.56 (0.62)  \\
      PointCloud & - & 1018872 & 23.64 (0.21)  \\ 
      Laplace & 16 & 1018872 & 17.34 (0.52) \\
      Dirac & 8 & 1018872 & {\bf 16.84 (0.16)}  \\
      \hline
    \end{tabular}
    \caption{Evaluation of different models on the temporal task}
    \label{tab:temporal}
    \vspace*{-2mm}
\end{table*}

\begin{figure*}
\centering
 \begin{tabular}{c c c c c} 
Ground Truth & MLP & PointCloud & Laplace & Dirac \\
 \vspace*{-6mm}
 \includegraphics[width=0.17\textwidth]{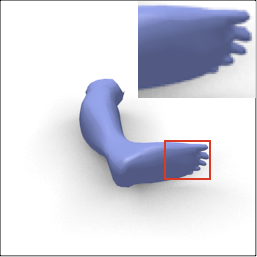} &
 \includegraphics[width=0.17\textwidth]{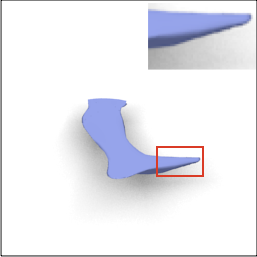} &
 \includegraphics[width=0.17\textwidth]{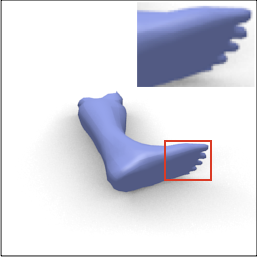} &
 \includegraphics[width=0.17\textwidth]{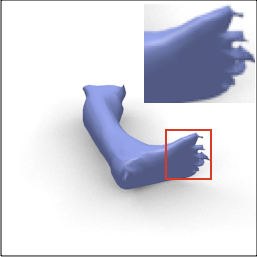} &
 \includegraphics[width=0.17\textwidth]{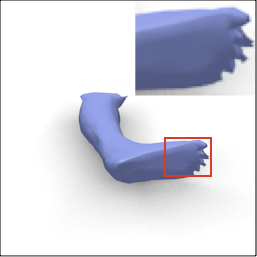} \\
  \includegraphics[width=0.17\textwidth]{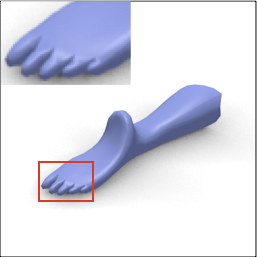} &
 \includegraphics[width=0.17\textwidth]{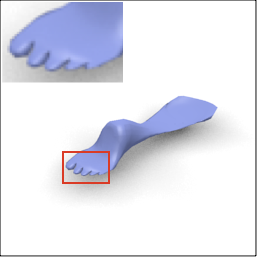} &
 \includegraphics[width=0.17\textwidth]{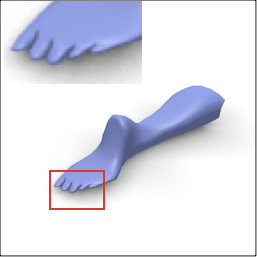} &
 \includegraphics[width=0.17\textwidth]{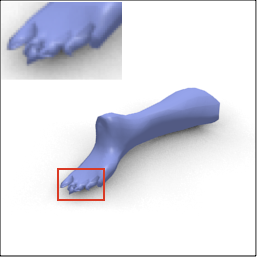} &
 \includegraphics[width=0.17\textwidth]{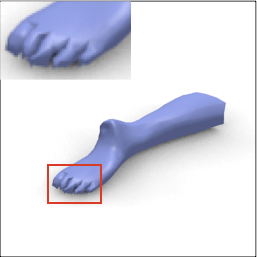} \\
 \end{tabular}
\caption{Qualitative comparison of different models. We plot 30th predicted frames correspondingly for two sequences in the test set. Boxes indicate distinctive features. For larger crops, see Figure \ref{fig:arap2}}.  
\vspace{-5mm}
 \label{fig:arap_img}
\end{figure*}

One specific task we consider is temporal predictions of non-linear dynamics. Given a sequence of frames $X = {X^1, X^2, \ldots, X^n}$, the task is to predict the following frames $Y = {Y^1=X^{n+1}, Y^2, \ldots, Y^m=X^{n+m}}$. As in \cite{mathieu2015deep}, we use a simple non-recurrent model that takes a concatenation of input frames $X$ and predicts a concatenation of frames $Y$. We condition on $n=2$ frames and predict the next $m=40$ frames.
In order to generate data, we first extracted 10k patches from the MPI-Faust dataset\cite{bogo2014faust}, by selecting a random point and growing a topological sphere of radius $15$ edges (i.e. the 15-ring of the point). For each patch, we generate a sequence of $50$ frames by randomly rotating it and letting it fall to the ground. We consider the mesh a thin elastic shell, and we simulate it using the As-Rigid-As-Possible technique \cite{sorkine2007rigid}, with additional gravitational forces \cite{Jacobson:Thesis}. Libigl \cite{libigl} has been used for the mesh processing tasks. Sequences with patches from the first 80 subjects were used in training, while the 20 last subjects were used for testing. 
The dataset and the code are available on request. 
We restrict our experiments to temporal prediction tasks that are deterministic when conditioned on several initial frames. Thus, we can train models by minimizing smooth-L1 loss \cite{girshick2015fast} between target frames and output of our models. 

We used models with 15 ResNet-v2 blocks with $128$ output features each. In order to cover larger context for Dirac and Laplace based models, we alternate these blocks with Average Pooling blocks. We predict offsets to the last conditioned frame and use the corresponding Laplace and Dirac operators. Thus, the models take 6-dimensional inputs and produce 120-dimensional outputs. We trained all models using the Adam optimizer \cite{kingma2015adam} with learning rate $10^{-3}$,  weight decay $10^{-5}$, and batch size 32. After 60k steps we decreased the learning rate by a factor of 2 every 10k steps. The models were trained for 110k steps in overall.

\begin{figure}[ht!]
\centering
 \begin{tabular}{c c c} 
 Ground Truth & Laplace & Dirac \\
 \includegraphics[width=0.3\columnwidth]{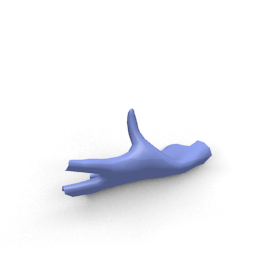} &
 \includegraphics[width=0.3\columnwidth]{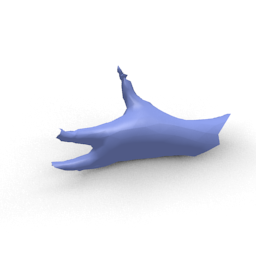} &
 \includegraphics[width=0.3\columnwidth]{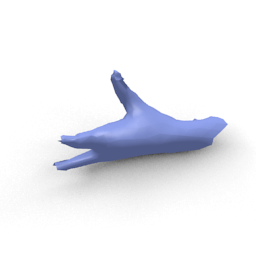} \\
 \includegraphics[width=0.3\columnwidth]{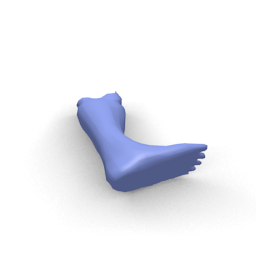} &
 \includegraphics[width=0.3\columnwidth]{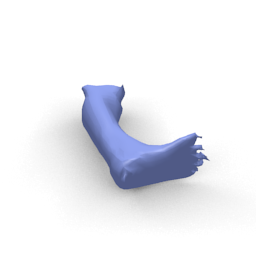} &
 \includegraphics[width=0.3\columnwidth]{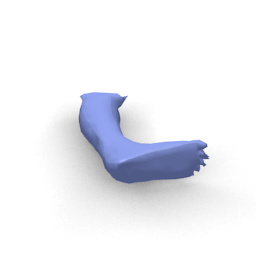}
 \end{tabular}
\caption{Dirac-based model visually outperforms Laplace-based models in the regions of high mean curvature.}  
\label{fig:arap2}
\end{figure}

\begin{figure}[ht!]
\centering
\includegraphics[width=0.45\textwidth,trim=0 0.25cm 0 0.75cm, clip]{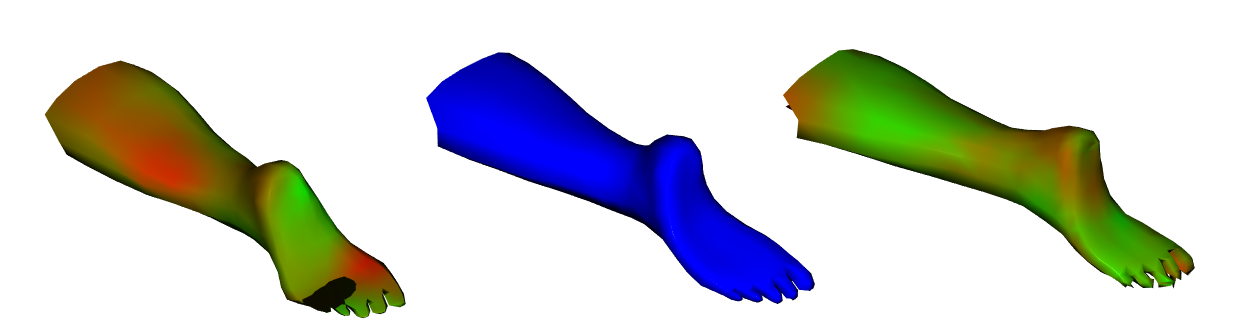} \\
\vspace{0.5cm}
\includegraphics[width=0.45\textwidth,trim=0 0.25cm 0 0.75cm, clip]{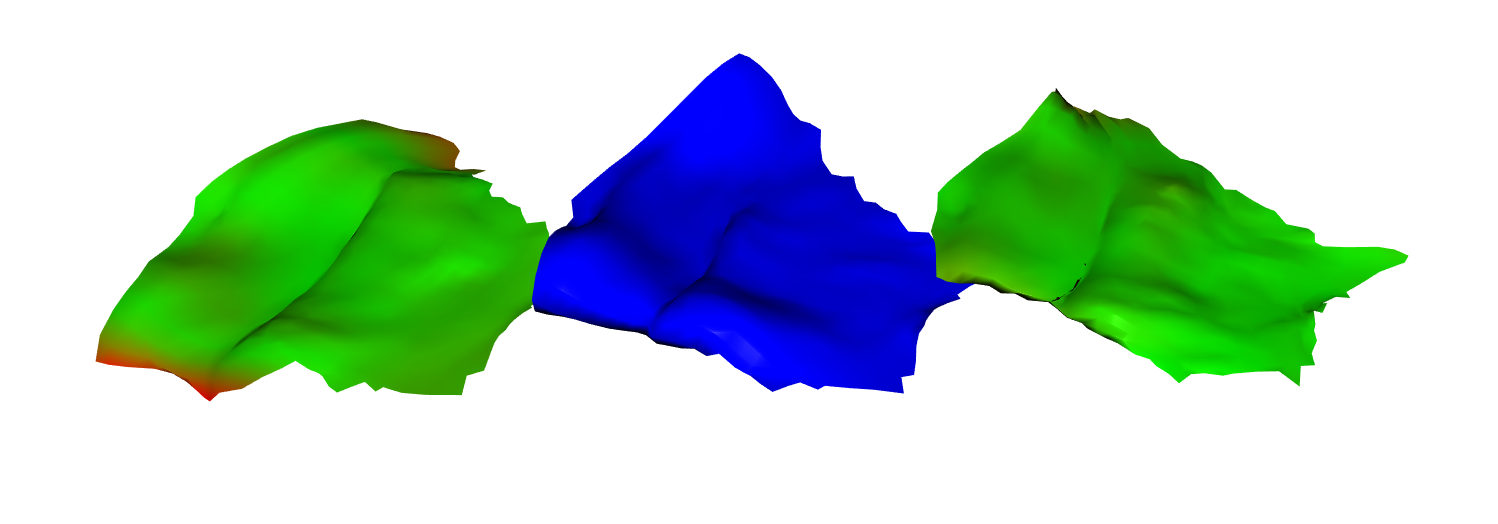} \\
\vspace{0.5cm}
\includegraphics[width=0.45\textwidth,trim=0 0.25cm 0 0.75cm, clip]{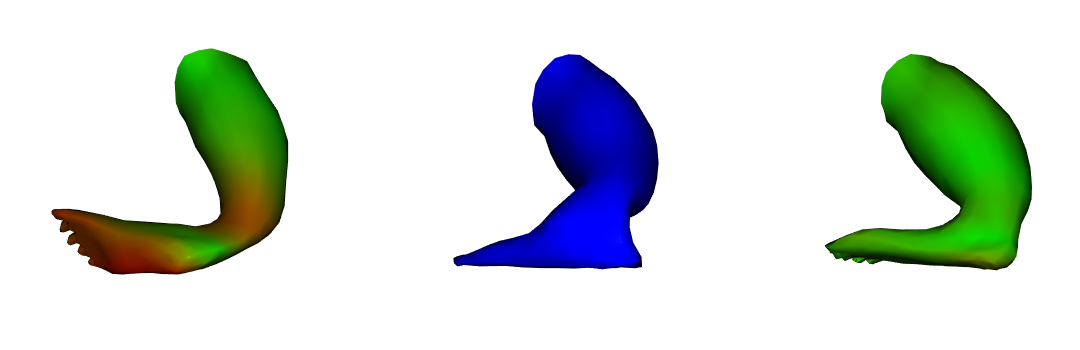}
\caption{From left to right: PointCloud (set2set), ground truth and Dirac based model. Color corresponds to mean squared error between ground truth and prediction: green - smaller error, red - larger error.\vspace{-5mm}}  
\label{fig:avgvsdir}
\end{figure}

\begin{figure}[ht!]
\centering
\includegraphics[width=0.4\textwidth,trim=0 0.25cm 0 0.75cm, clip]{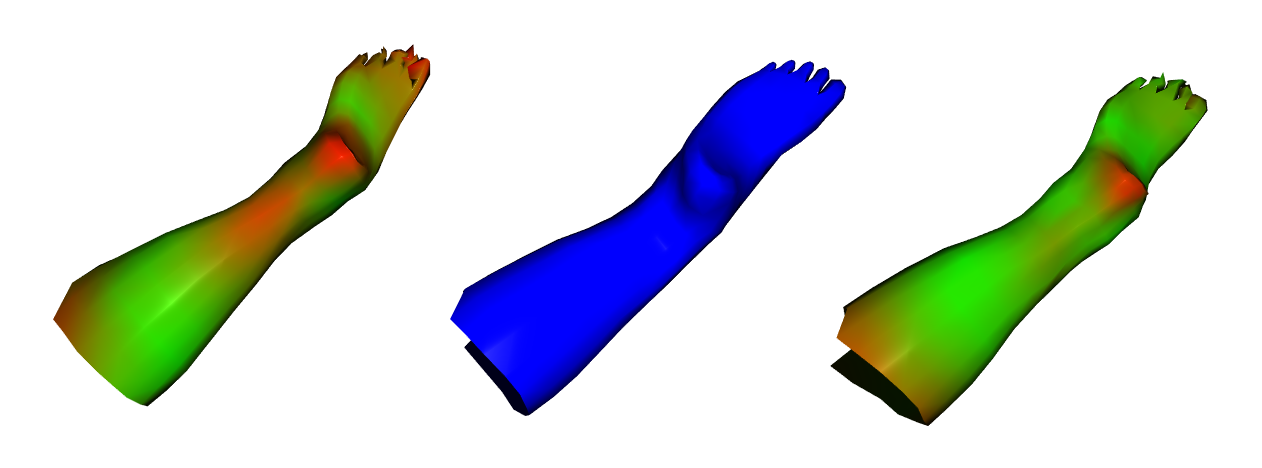} \\
\includegraphics[width=0.4\textwidth,trim=0 0.25cm 0 0.75cm, clip]{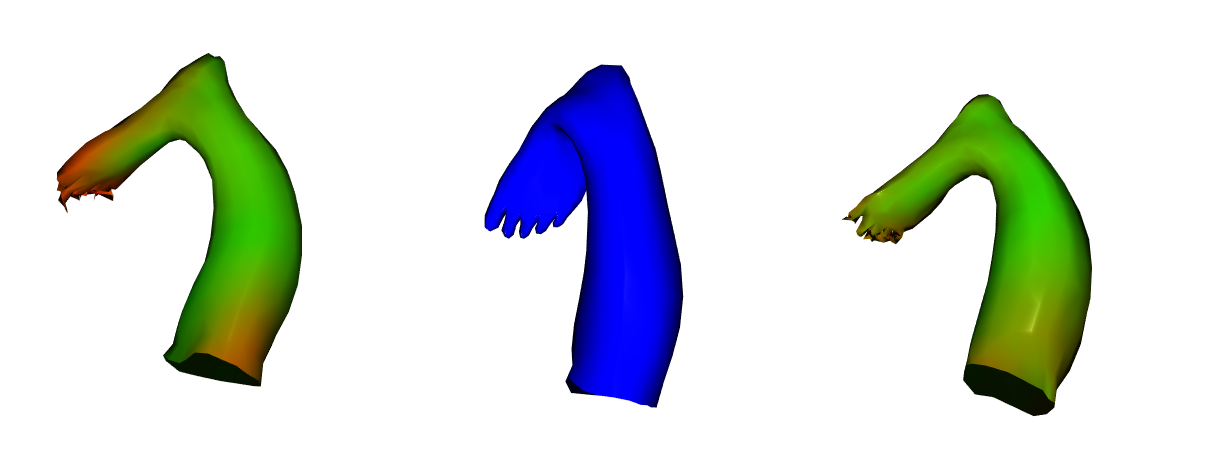} \\
\includegraphics[width=0.4\textwidth,trim=0 0.25cm 0 0.75cm, clip]{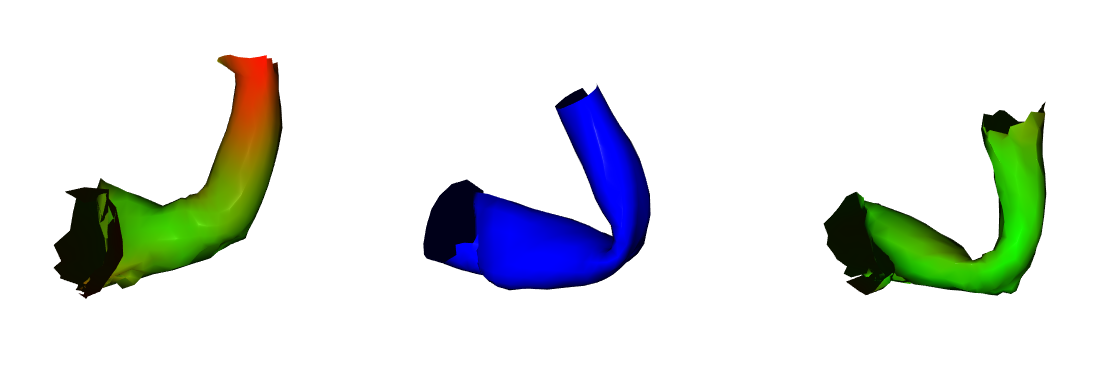}
\includegraphics[width=0.02\textwidth]{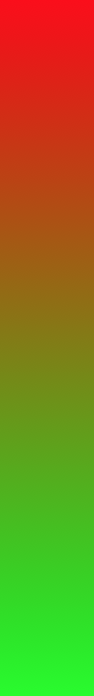}
\caption{From left to right: Laplace, ground truth and Dirac based model. Color corresponds to mean squared error between ground truth and prediction: green - smaller error, red - larger error.\vspace{-5mm}}  
\label{fig:lapvsdir}
\end{figure}

Table \ref{tab:temporal} reports quantitative prediction performance of different models, and Figure \ref{fig:arap_img} displays samples from the prediction models at specific frames. 
The set-to-set model \cite{vinyals2015order,sukhbaatar2016learning}, corresponding to a point-cloud representation used 
also in \cite{qi2016pointnet}, already performs reasonably well on the task, even if the visual difference is noticeable.
Nevertheless, the gap between this model and Laplace-/Dirac-based models is significant, both visually and quantitatively. 
Dirac-based model outperforms Laplace-based model despite the smaller receptive field. Videos comparing the performance of different models are available in the additional material. 

Figure \ref{fig:arap2} illustrates the effect of replacing Laplace by Dirac in the formulation of the SN. Laplacian-based models, since they propagate information using an isotropic operator, have more difficulties at resolving corners and pointy structures than the Dirac operator, that is sensitive to principal curvature directions. However, the capacity of Laplace models to exploit the extrinsic information only via the input coordinates is remarkable and more computationally efficient than the Dirac counterpart. 
Figures \ref{fig:avgvsdir} and \ref{fig:lapvsdir} overlay the prediction error and compare Laplace against Dirac and PointCloud against Dirac respectively. They confirm first that SNs outperform the point-cloud based model, which often produce excessive flattening and large deformations, and next that first-order Dirac operators help resolve areas with high directional curvature. 
We refer to the supplementary material for additional qualitative results.

 \section{Conclusions}

We have introduced Surface Networks, a deep neural network that is designed to naturally exploit the non-Euclidean geometry of surfaces. We have shown how a first-order differential operator (the Dirac operator) can detect and adapt to geometric features beyond the local mean  curvature, the limit of what Laplacian-based methods can exploit. This distinction is important in practice, since areas with high directional curvature are perceptually important, as shown in the experiments. That said, the Dirac operator comes at increased computational cost due to the quaternion calculus, and it would be interesting to instead \emph{learn} the operator, akin to recent Message-Passing NNs \cite{gilmer2017neural} and explore whether Dirac is recovered. 

Whenever the data contains good-quality meshes, our experiments demonstrate that using intrinsic geometry offers vastly superior performance to point-cloud based models. While there are not many such datasets currently available, we expect them to become common in the next years, as scanning and reconstruction technology advances and 3D sensors are integrated in consumer devices.
SNs provide efficient inference, with predictable runtime, which makes them appealing across many areas of computer graphics, where a fixed, per-frame cost is required to ensure a stable framerate, especially in VR applications. Our future plans include applying Surface Networks precisely to having automated, data-driven mesh processing, and generalizing the generative model to arbitrary meshes, which will require an appropriate multi-resolution pipeline. 

    


{\small
\bibliographystyle{ieee}
\bibliography{biblio}
}

  \appendix


  \section{The Dirac Operator}

\label{app:dirac}

The quaternions $\HH$ is an extension of complex numbers. A quaternion $q \in \HH$ can be represented in a form $q=a + bi + cj + dk$ where $a,b,c,d$ are real numbers and ${i,j,k}$ are quaternion units that satisfy the relationship $i^2 = j^2 = k^2 = ijk = -1$.

\begin{figure}[h]
\label{dir:triangle}
\centering
\begin{tikzpicture}
\filldraw (-1,0) circle(2pt) -- (1,0) circle(2pt) -- (0,1.44) circle(2pt);
\draw (-1,0)  node[anchor=north]{} -- (1,0) node[anchor=north]{ } -- (0,1.44) node[anchor=south]{$v_j$} -- cycle;
\draw[->, line width=1.25pt] (-1,-0.15) -- (0, -0.15) node[below]{$e_j$} -- (1,-0.15);
\draw[] (0.0, 0.67) node[]{f};
\end{tikzpicture}
\end{figure}

As mentioned in Section 3.1, the Dirac operator used in the model can be conveniently represented as a quaternion matrix:
$$D_{f,j} = \frac{-1}{2 | \ba_f | }e_j~,~f \in F, j \in V~,$$
where $e_j$ is the opposing edge vector of node $j$ in the face $f$, and $\ba_f$ is 
the area, as illustrated in Fig. \ref{dir:triangle}, 
using counter-clockwise orientations on all faces. 

The Deep Learning library PyTorch that we used to implement the models does not support quaternions. Nevertheless, quaternion-valued matrix multiplication can be replaced with real-valued matrix multiplication where each entry $q = a + bi + cj + dk$ is represented as a $4 \times 4$ block

$$
\begin{bmatrix}
    a & -b & -c & -d \\
    b & \phantom{-}a & -d &  \phantom{-}c \\
    c &  \phantom{-}d &  \phantom{-}a & -b \\
    d & -c &  \phantom{-}b &  \phantom{-}a
\end{bmatrix}
$$

and the conjugate $q^*=a-bi-cj-dk$ is a transpose of this real-valued matrix:

$$
\begin{bmatrix}
     \phantom{-}a &  \phantom{-}b &  \phantom{-}c &  \phantom{-}d \\
    -b &  \phantom{-}a &  \phantom{-}d & -c \\
    -c & -d &  \phantom{-}a &  \phantom{-}b \\
    -d &  \phantom{-}c & -b &  \phantom{-}a
\end{bmatrix}.
$$


  \section{Theorem 4.1}
 
 \subsection{Proof of (a)}
 We first show the result for the mapping $x \mapsto \rho \left( A x + B \Delta x \right)$, 
 corresponding to one layer of $\Phi_\Delta$. 
 By definition, the Laplacian $\Delta$ of $\M$ is 
 $$\Delta = \text{diag}(\bar{\ba})^{-1}( U - W)~,~$$
where $\bar{\ba}_j$ is one third of the total area of triangles incident to node $j$, 
and $W=(w_{i,j})$ contains the cotangent  weights \cite{wardetzky2008convergence}, and $U = \text{diag}(W {\bf 1})$ 
contains the node aggregated weights in its diagonal.

From \cite{das2007extremal} we verify that 
\begin{eqnarray}
\label{ble4}
\| U - W \| &\leq& \sqrt{2} \max_i \left\{\sqrt{ U_i^2 + U_i \sum_{i \sim j} U_j w_{i,j} } \right\} \\
&\leq& 2\sqrt{2} \sup_{i,j} w_{i,j} \sup_j d_j \nonumber \\
&\leq& 2 \sqrt{2} \cot ( \alpha_{\text{min}} ) d_{\text{max}} ~, \nonumber
\end{eqnarray}
where $d_j$ denotes the degree (number of neighbors) of node $j$, 
$\alpha_{\text{min}}$ is the smallest angle in the triangulation of $\M$ and $S_{\text{max}}$ the 
largest number of incident triangles. 
It results that 
$$\| \Delta \| \leq C \frac{\cot ( \alpha_{\text{min}} ) S_{\text{max}}}{\inf_j \bar{\ba}_j}:= L_\M~,$$
which depends uniquely on the mesh $\M$ and is finite for non-degenerate meshes. 
Moreover, since $\rho(\,\cdot \,)$ is non-expansive, we have
\begin{eqnarray}
\label{za1}
\left\| \rho \left( A x + B \Delta x \right) - \rho\left( A x' + B \Delta x' \right) \right \| & \leq & \| A( x - x') + B \Delta (x-x') \| \\ \nonumber
& \leq & (\| A \| + \| B \| L_\M ) \| x - x'\|~. 
\end{eqnarray}

By cascading (\ref{za1}) across the $K$ layers of the network, we obtain
\begin{equation*}
\| \Phi(\M; x) - \Phi(\M; x') \| \leq \left(\prod_{k \leq K} ( \| A_k \| + \| B_k \| L_\M) \right) \| x - x' \|~,
\end{equation*}
which proves (a). $\square$

\subsection{Proof of (b)}

The proof is analogous, by observing that $\| D \| = \sqrt{ \| \Delta \| }$ and therefore 
$$\| D \| \leq  \sqrt{L_\M}~.~~~\square$$

\subsection{Proof of (c)}

To establish (c) we first observe that given three points $p, q, r \in \R^3$ forming any of the triangles of $\M$, 
{\small 
\begin{eqnarray}
\| p - q \|^2 (1 - \taunorm)^2 &\leq \| \tau(p) - \tau(q) \|^2 \leq& \| p - q \|^2 (1 + \taunorm)^2 \label{za3} \\
\ba(p,q,r)^2 ( 1 - \taunorm C \alpha_{\text{min}}^{-2} - o(\taunorm^2) & \leq \ba(\tau(p), \tau(q), \tau(r) )^2 \leq & \ba(p,q,r)^2 ( 1 + \taunorm C \alpha_{\text{min}}^{-2} + o(\taunorm^2))~.\label{za44}   
\end{eqnarray}}
Indeed, (\ref{za3}) is a direct consequence of the lower and upper Lipschitz constants of $\tau(u)$, which are bounded respectively by $1- \taunorm$ 
and $1 + \taunorm$. As for (\ref{za44}), we use the Heron formula 
$$\ba(p,q,r)^2 = s ( s - \| p - q \|)( s - \| p - r \|)( s - \| r - q \|)~,$$
with $s = \frac{1}{2}( \| p - q \| + \| p - r \| + \| r - q \|)$ being the half-perimeter.
By denoting $s_\tau$ the corresponding half-perimeter determined by the deformed points $\tau(p), \tau(q), \tau(r)$, 
we have that 
$$ s_\tau - \| \tau(p) - \tau(q) \| \leq s(1 + \taunorm) - \| p - q \| ( 1 - \taunorm) = s -  \| p - q \|  + \taunorm( s + \| p - q \|)~\text{and }$$
$$ s_\tau - \| \tau(p) - \tau(q) \| \geq s(1 - \taunorm) - \| p - q \| ( 1 + \taunorm) = s -  \| p - q \|  - \taunorm( s + \| p - q \|)~,$$
and similarly for the $\| r - q\|$ and $\|r - p \|$ terms. 
It results in
\begin{eqnarray*}
\ba(\tau(p),\tau(q),\tau(r))^2 &\geq& \ba(p,q,r)^2 \left[ 1 - \taunorm \left( 1 + \frac{s + \| p - q \|}{s - \| p -q \|} + \frac{s + \| p - r \|}{s - \| p -r \|} + \frac{s + \| r - q \|}{s - \| r -q \|} \right) - o( \taunorm^2 )\right ]  \\
&\geq & \ba(p,q,r)^2 \left[ 1 - C \taunorm \alpha_{\text{min}}^{-2}  - o( \taunorm^2) \right]~,
\end{eqnarray*}
and similarly 
$$\ba(\tau(p),\tau(q),\tau(r))^2  \leq \ba(p,q,r)^2 \left[ 1 + C \taunorm \alpha_{\text{min}}^{-2}  - o( \taunorm^2) \right] ~.$$

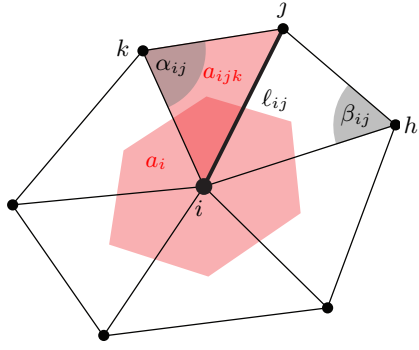
\begin{figure}
\centering
\begin{overpic}
[width=0.3\columnwidth]{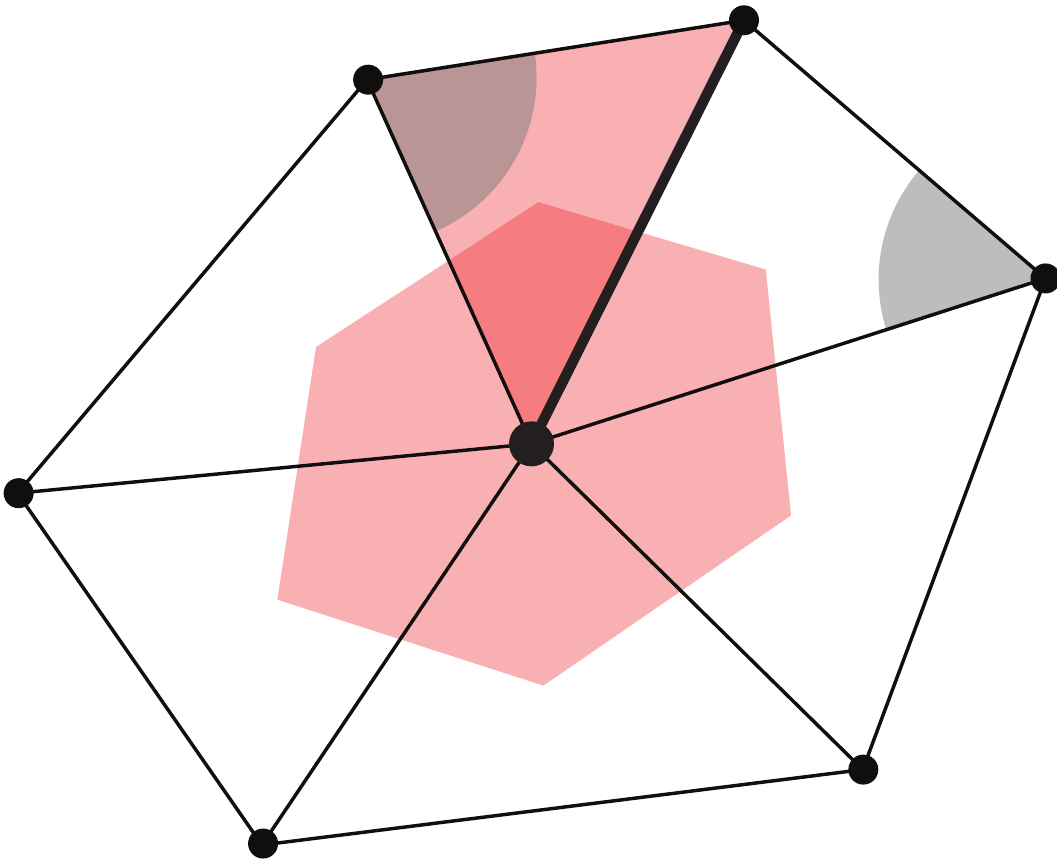}
	\put(69,83.5){\small $j$} 		
	\put(48,32){\small $i$} 		
	\put(28,72.5){\small$k$} 	
	\put(101,53){\small$h$} 	
	\put(38,69){\small$\alpha_{ij}$} 	
	\put(85,56){\small$\beta_{ij}$} 	
	\put(35.5,44.5){\color{red}\small$a_i$} 	
	\put(50,67){\color{red}\small$a_{ijk}$} 	
	\put(65,60){\small$\ell_{ij}$} 	
	\end{overpic}
\caption{Triangular mesh and Cotangent Laplacian (figure reproduced from \cite{bronstein2016geometric}) }
\label{figcotangent}
\end{figure}

By noting that the cotangent Laplacian weights can be written (see Fig. \ref{figcotangent}) as
$$w_{i,j} = \frac{- \ell_{ij}^2 + \ell_{jk}^2 + \ell_{ik}^2 }{\ba(i,j,k)} + \frac{- \ell_{ij}^2 + \ell_{jh}^2 + \ell_{ih}^2 }{\ba(i,j,h)}~, $$
we have from the previous Bilipschitz bounds that
$$\tau( w_{i,j}) \leq w_{i,j} \left[ 1 - C \taunorm \alpha_{\text{min}}^{-2}\right]^{-1} + 2 \taunorm \left[ 1 - C \taunorm \alpha_{\text{min}}^{-2}\right]^{-1} \left( \frac{\ell_{ij}^2 + \ell_{jk}^2 + \ell_{ik}^2}{\ba(i,j,k)} + \frac{\ell_{ij}^2 + \ell_{jh}^2 + \ell_{ih}^2}{\ba(i,j,h)} \right)~,$$
$$\tau( w_{i,j}) \geq w_{i,j} \left[ 1 + C \taunorm \alpha_{\text{min}}^{-2}\right]^{-1} - 2 \taunorm \left[ 1 + C \taunorm \alpha_{\text{min}}^{-2}\right]^{-1} \left( \frac{\ell_{ij}^2 + \ell_{jk}^2 + \ell_{ik}^2}{\ba(i,j,k)} + \frac{\ell_{ij}^2 + \ell_{jh}^2 + \ell_{ih}^2}{\ba(i,j,h)} \right)~,$$
which proves that, up to second order terms, the cotangent weights are Lipschitz continuous to deformations. 

Finally, since the mesh Laplacian operator is constructed as $\text{diag}(\bar{\ba})^{-1} (U - W)$, 
with $\bar{\ba}_{i,i} = \frac{1}{3} \sum_{j,k; (i,j,k) \in F} \ba(i,j,k)$, and $U = \text{diag}( W {\bf 1})$,
let us show how to bound $\| \Delta - \tau(\Delta) \|$ from
\begin{equation}
\label{pep1}
\bar{\ba}_{i,i} ( 1 - \alpha_\M \taunorm - o( \taunorm^2) ) \leq \tau(\bar{\ba}_{i,i}) \leq \bar{\ba}_{i,i} ( 1 + \alpha_\M \taunorm + o( \taunorm^2) )
\end{equation} 
and
\begin{equation}
\label{pep2}
w_{i,j} ( 1 - \beta_\M \taunorm - o( \taunorm^2) ) \leq \tau(w_{i,j}) \leq w_{i,j} ( 1 + \beta_\M \taunorm + o( \taunorm^2) )~.
\end{equation} 
Using the fact that $\bar{\ba}$, $\tau(\bar{\ba})$ are diagonal, and using the spectral bound for $k \times m$ sparse matrices 
from \cite{chen2005obtaining}, Lemma 5.12, 
$$\| Y \|^2 \leq \max_i \sum_{j ; \, Y_{i,j} \neq 0} |Y_{i,j}| \left( \sum_{r=1}^l | Y_{r,j}| \right)~, $$
the bounds (\ref{pep1}) and (\ref{pep2}) 
yield respectively 
\begin{eqnarray*}
\label{pep3}
 \tau(\bar{\ba}) &=& \bar{\ba} ( {\bf 1} + \epsilon_\tau)~,~\text{with } \| \epsilon_\tau \| = o( \taunorm)~,\text{and}  \\
 \tau( U - W) &=& U - W + \eta_\tau~,~\text{with } \| \eta_\tau \| = o ( \taunorm )~.
\end{eqnarray*}
It results that, up to second order terms, 
\begin{eqnarray*}
\| \Delta - \tau(\Delta) \| &=& \left \| \tau(\bar{\ba})^{-1} ( \tau(U) - \tau(W) ) - \bar{\ba}^{-1} ( U - W) \right\| \\
&=& \left \| \left( \bar{\ba} [{\bf 1} + \epsilon_\tau ] \right)^{-1} \left[ U - W + \eta_\tau \right] - \bar{\ba}^{-1} ( U - W) \right\| \\
&=& \left \| \left( {\bf 1} - \epsilon_\tau + o(\taunorm^2) \right) \bar{\ba}^{-1} ( U - W + \eta_\tau) - \bar{\ba}^{-1} ( U - W)  \right\| \\
&=& \left \| \epsilon_\tau \Delta + \bar{\ba}^{-1} \eta_\tau \right\| + o( \taunorm^2) \\ 
&=& o( | \tau|_\infty)~,
\end{eqnarray*}
which shows that the Laplacian is stable to deformations in operator norm. 
Finally, by denoting $\tilde{x}_\tau$ a layer of the deformed Laplacian network 
$$\tilde{x}_\tau = \rho( A x + B \tau(\Delta) x)~,$$
it follows that 
\begin{eqnarray}
\label{pep4}
\| \tilde{x} - \tilde{x}_\tau \| &\leq& \| B ( \Delta - \tau(\Delta) x \| \\
&\leq & C \| B \|\taunorm \|x \| ~.
\end{eqnarray}
Also, 
\begin{eqnarray}
\label{pep42}
\| \tilde{x} - \tilde{y}_\tau \| &\leq& \| A( x - y) + B ( \Delta x - \tau(\Delta) y) \| \nonumber \\
&\leq & (\| A \| + \| B \| \| \Delta \|  )\| x - y\|  + \| \Delta - \tau(\Delta) \| \| x\|  \nonumber \\
& \leq & \underbrace{(\| A \| + \| B \| \| \Delta \|  )}_{\delta_1}\| x - y\|  + \underbrace{C \taunorm}_{\delta_2} \| x \|~, 
\end{eqnarray}
and therefore, by plugging (\ref{pep42}) with $y = \tilde{x}_\tau$, 
$K$ layers of the Laplacian network satisfy 
\begin{eqnarray*}
\| \Phi(x; \Delta) - \Phi(x; \tau(\Delta) \| &\leq& \left(\prod_{j \leq K-1} \delta_1(j)\right) \| \tilde{x} - \tilde{x}_\tau\| + \left(\sum_{j < K-1} \prod_{j' \leq j} \delta_1(j') \delta_2(j) \right)  \taunorm \|x \| \\
&\leq & \left[C  \left(\prod_{j \leq K-1} \delta_1(j)\right)  \|B \| +  \left(\sum_{j < K-1} \prod_{j' \leq j} \delta_1(j') \delta_2(j) \right)  \right] \taunorm \|x \| ~. ~~~ \square~.
\end{eqnarray*}

\subsection{Proof of (d)}

The proof is also analogous to the proof of (c), with the difference that now 
the Dirac operator is no longer invariant to orthogonal transformations, only to translations.
Given two points $p$, $q$, we verify that 
$$\| p - q - \tau(p) - \tau(q) \| \leq \widetilde{| \tau |}_\infty \| p - q \|~,$$
which, following the previous argument, leads to 
\begin{equation}
\| D - \tau(D) \| = o( \widetilde{| \tau |}_\infty )~.
\end{equation}

\section{Theorem 4.2}

\subsection{Proof of part (a)}

The proof is based on the following lemma:
\begin{lemma}
\label{lemma1}
Let $x_N,y_N \in \mathcal{H}(\M_N)$ such that $\forall ~N$, $\| x_N \|_{\Hc} \leq c$,$\| y_N \|_{\Hc} \leq c$. Let $\hat{x_N} = \mathcal{E}_N(x_N)$, where $\mathcal{E}_N$ is the eigendecomposition of the Laplacian operator $\Delta_N$ on $\M_N$, , with 
associated eigenvalues $\lambda_1 \dots \lambda_N$ in increasing order. 
Let $\gamma>0$ and $\beta$ be defined as in (\ref{betadef}) for $x_N$ and $y_N$. 
If $\beta > 1$ and $\|x_N - y_N\| \leq \epsilon$ for all $N$, 
\begin{equation}
\label{gu1}
\| \Delta_N(x_N - y_N) \|^2 \leq C \epsilon^{2 - \frac{1}{\beta-1/2}}~,
\end{equation}
where $C$ is a constant independent of $\epsilon$ and $N$.
\end{lemma}

One layer of the network will transform the difference $x_1 - x_2$ into 
$\rho( A x_1 + B \Delta x_1) - \rho( A x_2 + B \Delta x_2)$. We verify that
\begin{equation*}
\| \rho( A x_1 + B \Delta x_1) - \rho( A x_2 + B \Delta x_2) \| \leq  \| A \| \| x_1 - x_2 \| + \| B \| \| \Delta (x_1 - x_2) \|~.
\end{equation*}
We now apply Lemma \ref{lemma1} to obtain 
\begin{eqnarray*}
\| \rho( A x_1 + B \Delta x_1) - \rho( A x_2 + B \Delta x_2) \| &\leq& \|A \| \| x_1 - x_2 \| + C \| B \| \| x_1 - x_2 \|^{\frac{\beta-1}{\beta-1/2}} \nonumber \\
&\leq& \| x_1 - x_2 \|^{\frac{\beta-1}{\beta-1/2}} \left( \|A \| \| x_1 - x_2 \|^{(2\beta-1)^{-1}} + C \| B \| \right) \nonumber  \\
&\leq & C ( \|A \| + \|B \|) \| x_1 - x_2 \|^{\frac{\beta-1}{\beta-1/2}}~,
\end{eqnarray*}
where we redefine $C$ to account for the fact that $\| x_1 - x_2 \|^{(2\beta-1)^{-1}}$ is bounded. 
We have just showed that
\begin{equation}
\label{savo1}
\| x^{(r+1)}_1 - x^{(r+1)}_2 \| \leq f_r \| x^{(r)}_1 - x^{(r)}_2 \|^{g_r}
\end{equation}
with $f_r = C ( \|A_r \| + \|B_r \|)$ and $g_r =\frac{\beta_r-1}{\beta_r-1/2}$. 
By cascading (\ref{savo1}) for each of the $R$ layers we thus obtain
\begin{equation}
\| \Phi_\Delta(x_1) - \Phi_\Delta(x_2) \| \leq \left[\prod_{r = 1}^R f_r^{\prod_{r'>r} g_{r'}} \right] \| x_1 - x_2 \|^{\prod_{r=1}^R g_{r}}~,
\end{equation}
which proves (\ref{pony1})   $\square$.

{\it Proof of (\ref{gu1}): } 
Let $\{e_1, \dots, e_N\}$ be the eigendecomposition of $\Delta_N$. 
For simplicity, we drop the subindex $N$ in the signals from now on.
Let $\hat{x}(k) = \langle x, e_k \rangle $ and $\tilde{x}(k) = \lambda_k \hat{x}(k)$; and analogously for $y$. From the Parseval identity we have that $\| x\|^2 = \| \hat{x} \|^2$.
We express $\| \Delta(x - y) \|$ as
\begin{equation}
\label{bv1}
\| \Delta(x - y) \|^2 = \sum_{k\leq N} \lambda_k^2 ( \hat{x}(k) - \hat{y}(k))^2~.
\end{equation}
The basic principle of the proof is to cut the spectral sum (\ref{bv1}) in two parts, 
chosen to exploit the decay of $\tilde{x}(k)$. Let
$$F(x)(k) = \frac{\sum_{k' \geq k} \tilde{x}(k)^2}{\| x \|_{\Hc}^2} = \frac{\sum_{k' \geq k} \tilde{x}(k)^2}{\sum_{k'} \tilde{x}(k)^2} = \frac{\sum_{k' \geq k} \lambda_k^2 \hat{x}(k)^2}{\sum_{k'}\lambda_k^2 \hat{x}(k)^2} \leq 1~, $$
and analogously for $y$. 
For any cutoff $k_* \leq N$ we have
\begin{eqnarray}
\label{bv3}
\| \Delta(x - y) \|^2 &=& \sum_{k \leq k_*} \lambda_k^2 ( \hat{x}(k) - \hat{y}(k))^2 + \sum_{k > k_*} \lambda_k^2 ( \hat{x}(k) - \hat{y}(k))^2 \nonumber \\
&\leq & \lambda_{k_*}^2 \epsilon^2 + 2(F(x)(k_*) \| x \|_{\Hc}^2 + F(y)(k_*) \| y \|_{\Hc}^2) \nonumber \\
&\leq & \lambda_{k_*}^2 \epsilon^2 + 2F(k_*) ( \| x \|_{\Hc}^2 + \| y \|_{\Hc}^2) \nonumber \\
& \leq & \lambda_{k_*}^2 \epsilon^2 + 4F(k_*) D^2~,
\end{eqnarray}
where we denote for simplicity $F(k_*) = \max(F(x)(k_*),~F(y)(k_*))$. 
By assumption, we have $\lambda_k^2 \lesssim k^{2\gamma}$ 
and 
$$F(k) \lesssim \sum_{k' \geq k} k^{2(\gamma - \beta)} \simeq k^{1 + 2(\gamma - \beta)}~.$$
By denoting $\tilde{\beta} = \beta - \gamma -1/2$, it follows that
\begin{equation}
\label{bv4}
\| \Delta(x - y) \|^2 \lesssim \epsilon^2 k_*^{2\gamma} + 4D^2 k_*^{-2\tilde{\beta}}   
\end{equation}
Optimizing for $k_*$ yields 
$$\epsilon^2 2\gamma k^{2\gamma-1} -2\tilde{\beta} 4D^2 k^{-2\tilde{\beta}-1} = 0, \text{ thus }$$
\begin{equation}
\label{bv2}
k_* = \left[\frac{4 \beta D^2}{\gamma \epsilon^2} \right]^{\frac{1}{2\gamma + 2\tilde{\beta}}}~.
\end{equation}
By plugging (\ref{bv2}) back into (\ref{bv4}) and dropping all constants independent of $N$ and $\epsilon$, this leads to 
$$\| \Delta(x - y) \|^2 \lesssim \epsilon^{2 - \frac{1}{\gamma + \tilde{\beta} }} = \epsilon^{2 - \frac{1}{\beta-1/2}}~,$$
which proves part (a) $\square$.


\subsection{Proof of part (b)}

We will use the following lemma:
\begin{lemma}
\label{lemma2}
Let $\M=(V, E, F)$ is a non-degenerate mesh, and define 
\begin{equation}
\eta_1(\M) = \sup_{(i,j) \in E} \frac{\bar{\ba}_i }{\bar{\ba}_j}~,~\eta_2(\M) = \sup_{(i,j,k) \in F} \frac{\ell_{ij}^2 + \ell_{jk}^2 + \ell_{ik}^2}{\ba(i,j,k)}~,~\eta_3(\M) = \alpha_{\text{min}}~.
\end{equation}
Then, given a smooth deformation $\tau$ and $x$ defined in $\M$, we have
\begin{equation}
\label{ui1}
\| (\Delta - \tau(\Delta)) x \| \leq C \taunorm \| \Delta x\|~,
\end{equation}
where $C$ depends only upon $\eta_1$, $\eta_2$ and $\eta_3$.  
\end{lemma}

In that case, we need to control the difference  
$\rho( A x + B \Delta x) - \rho( A x + B \tau(\Delta) x)$. We verify that
\begin{equation*}
\| \rho( A x + B \Delta x) - \rho( A x + B \tau(\Delta) x) \| \leq   \| B \| \| (\Delta - \tau(\Delta)) x \|~.
\end{equation*}
By Lemma \ref{lemma2} it follows that $\| (\Delta - \tau(\Delta)) x \| \leq C \taunorm \| \Delta x\|$ 
and therefore, by denoting $x_1^{(1)} = \rho( A x + B \Delta x)$ and 
$x_2^{(1)} = \rho( A x + B \tau(\Delta) x)$, we have
\begin{equation}
\label{vic1}
\| x_1^{(1)} - x_2^{(1)} \| \leq C \taunorm \| \Delta x\| = C \taunorm \| x \|_{\Hc}~.
\end{equation}
By applying again Lemma \ref{lemma1}, we also have that 
\begin{eqnarray*}
\| \Delta x_1^{(1)} - \tau(\Delta) x_2^{(1)} \| &=& \| \Delta x_1^{(1)} - ( \Delta + \tau(\Delta) -\Delta)x_2^{(1)} \| \\
&=& \| \Delta ( x_1^{(1)} - x_2^{(1)}) + ( \tau(\Delta) -\Delta) x_2^{(1)} \| \\
&\leq& C \| x_1^{(1)} - x_2^{(1)} \|^{\frac{\beta_1-1}{\beta_1-1/2}} + \taunorm \| x_2^{(1)} \|_{\Hc} \\
&\lesssim & C \taunorm^{\frac{\beta_1-1}{\beta_1-1/2}} ~,
\end{eqnarray*}
which, by combining it with (\ref{vic1}) and repeating through the $R$ layers yields
\begin{equation}
\| \Phi_\Delta(x, \M) - \Phi_\Delta(x, \tau(\M) \| \leq C \taunorm^{\prod_{r=1}^R \frac{\beta_r-1}{\beta_r-1/2} }~,
\end{equation}
which concludes the proof  $\square$. 

{\it Proof of (\ref{ui1}):} 
The proof follows closely the proof of Theorem \ref{stabtheo}, part (c). 
From (\ref{pep1}) and (\ref{pep2}) we have that 
\begin{eqnarray*}
\label{seb3}
 \tau(\bar{\ba}) &=& \bar{\ba} ( {\bf I} + G_\tau)~,~\text{with } | G_\tau |_\infty \leq C(\eta_2, \eta_3)\taunorm~,\text{and}  \\
 \tau( U - W) &=& ({\bf I} + H_\tau)(U - W)~,~\text{with } | H_\tau |_\infty \leq C(\eta_2, \eta_3) \taunorm~.
\end{eqnarray*}
It follows that, up to second order $o(\taunorm^2)$ terms, 
\begin{eqnarray}
\label{seb4}
  \tau(\Delta) - \Delta &=& \tau(\bar{\ba})^{-1} ( \tau(U) - \tau(W) ) - \bar{\ba}^{-1} ( U - W) \nonumber \\
&=&  \left( \bar{\ba} [{\bf 1} + G_\tau ] \right)^{-1} \left[({\bf I} + H_\tau) (U - W) \right] - \bar{\ba}^{-1} ( U - W) \nonumber \\
&\simeq&  \bar{\ba}^{-1} H_\tau (U - W) + G_\tau \Delta ~.
\end{eqnarray}
By writing $\bar{\ba}^{-1} H_\tau = \widetilde{H_\tau} \bar{\ba}^{-1}$, 
and since $\bar{\ba}$ is diagonal, we verify that
$$(\widetilde{H_\tau})_{i,j} = (H_\tau)_{i,j} \frac{\ba_{i,i}}{\ba_{j,j}}~, \text{with } $$
$\frac{\ba_{i,i}}{\ba_{j,j}} \leq \eta_1$, and hence that 
\begin{equation}
\label{seb5}
\bar{\ba}^{-1} H_\tau (U - W) = \widetilde{H_\tau} \Delta~,~\text{with } |\widetilde{H_\tau}|_\infty \leq C(\eta_1, \eta_2, \eta_3) \taunorm~.
\end{equation}
We conclude by combining (\ref{seb4}) and (\ref{seb5}) into
\begin{eqnarray*}
\| ( \Delta - \tau(\Delta) ) x \| &=& \| ( G_\tau + \widetilde{H_\tau} ) \Delta x \| \\
&\leq& C'(\eta_1, \eta_2, \eta_3) \taunorm \| \Delta x \|~,
\end{eqnarray*}
which proves (\ref{ui1}) $\square$

\subsection{Proof of part (c)}

This result is a consequence of the consistency of the cotangent Laplacian to the Laplace-Beltrami operator on $S$ \cite{wardetzky2008convergence}:
\begin{theorem}[\cite{wardetzky2008convergence}, Thm 3.4] Let $\M$ be a compact polyhedral surface which is a normal graph over a smooth surface $S$ 
with distortion tensor $\mathcal{T}$, and let $\bar{\mathcal{T}} = (\det \mathcal{T})^{1/2} \mathcal{T}^{-1}$. 
If the normal field uniform distance $d(\mathcal{T}, {\bf 1} ) = \| \bar{\mathcal{T}} - {\bf 1} \|_\infty$ satisfies  $d(\mathcal{T}, {\bf 1}) \leq \epsilon$, then
\begin{equation}
\label{blabla1}
\| \Delta_\M - \Delta_S\| \leq \epsilon~.
\end{equation}
\end{theorem}

If $\Delta_\M$ converges uniformly to $\Delta_S$, 
in particular we verify that
$$\| x \|_{\mathcal{H}(\M)} \to \| x \|_{\mathcal{H}(S)}~.$$


Thus, given two meshes $\M$, $\M'$ approximating a smooth surface $S$ in terms of uniform normal distance, 
and the corresponding irregular sampling $x$ and $x'$ of an underlying function $\bar{x} : S \to \R$, we have 
\begin{equation}
\label{blabla2}
\| \rho( A x + B \Delta_{\M} x) - \rho( A x' + B \Delta_{\M'} x') \| \leq \| A \| \| x - x' \| + \|B \| \| \Delta_\M x - \Delta_{\M'} x' \|~.
\end{equation}
Since $\M$ and $\M'$ both converge uniformly normally to $S$ and $\bar{x}$ is Lipschitz on $S$, it results 
that 
$$\| x - \bar{x} \| \leq L \epsilon~,\text{ and }~\| x' - \bar{x} \| \leq L \epsilon~,$$
thus $\| x - x' \| \leq 2 L \epsilon$. 
Also, thanks to the uniform normal convergence, we also have convergence in the Sobolev sense:
$$\| x - \bar{x} \|_{\mathcal{H}} \lesssim \epsilon~,~\| x' - \bar{x} \|_{\mathcal{H}} \lesssim \epsilon~, $$
which implies in particular that 
\begin{equation}
\label{blabla3}
\| x - x' \|_{\mathcal{H}} \lesssim \epsilon~.
\end{equation}
From (\ref{blabla2}) and (\ref{blabla3}) it follows that
\begin{eqnarray}
\label{blabla4}
\| \rho( A x + B \Delta_{\M} x) - \rho( A x' + B \Delta_{\M'} x') \| &\leq& 2 \| A \| L \epsilon +  \\
&& + \| B \| \| \Delta_\M x - \Delta_{S} \bar{x} + \Delta_{S} \bar{x} - \Delta_{\M'} x'  \| \nonumber \\
&\leq & 2\epsilon \left(  \|A \| L + \| B \|  \right)  \nonumber ~.
\end{eqnarray}
By applying again Lemma \ref{lemma1} to $\tilde{x} = \rho( A x + B \Delta_{\M} x)$, $\tilde{x}'=\rho( A x' + B \Delta_{\M'} x')$, we have 
$$\| \tilde{x} - \tilde{x}' \|_{\mathcal{H}}  \leq C \| \tilde{x} - \tilde{x}'\|^{\frac{\beta_1-1}{\beta_1-1/2}} \lesssim \epsilon^{\frac{\beta_1-1}{\beta_1-1/2}}~.$$
We conclude by retracing the same argument as before, reapplying Lemma \ref{lemma1} 
at each layer to obtain
\begin{eqnarray*}
\| \Phi_\M(x) - \Phi_{\M'}(x') \| &\leq& C \epsilon^{\prod_{r=1}^R \frac{\beta_r-1}{\beta_r-1/2}}~.~~~\square~.
\end{eqnarray*}



\section{Proof of Corollary 4.3}

We verify that 
\begin{eqnarray*}
\| \rho(  B \Delta x) - \rho(  B \tau(\Delta) \tau(x) ) \| &\leq & \| B \| \| \Delta x - \tau(\Delta) \tau(x) \|  \\
&\leq& \| B \|  \| \Delta( x - \tau(x))  + (\Delta- \tau(\Delta))(\tau(x)) \| \\
&\leq & \| B \| ( \| \Delta( x - \tau(x)) \| + \| (\Delta- \tau(\Delta))(\tau(x)) \| ~.
\end{eqnarray*}
The second term is $o(\taunorm)$ from Lemma \ref{lemma2}. The first term is 
$$\| x - \tau(x) \|_{\Hc} \leq \| \Delta ( {\bf I} - \tau) \| \|x \| \leq \|\nabla^2 \tau\| \|x\|~,$$ 
where $\|\nabla^2 \tau\|$ is the uniform Hessian norm of $\tau$. 
The result follows from applying the cascading argument from last section. $\square$


  \section{Preliminary Study: Metric Learning for Dense Correspondence}
\label{densecorrespondencesect}
As an interesting extension, we apply the architecture we built in Experiments 6.2 directly to a dense shape correspondence problem.

\begin{figure}[h!]
\centering
    \includegraphics[width=\textwidth]{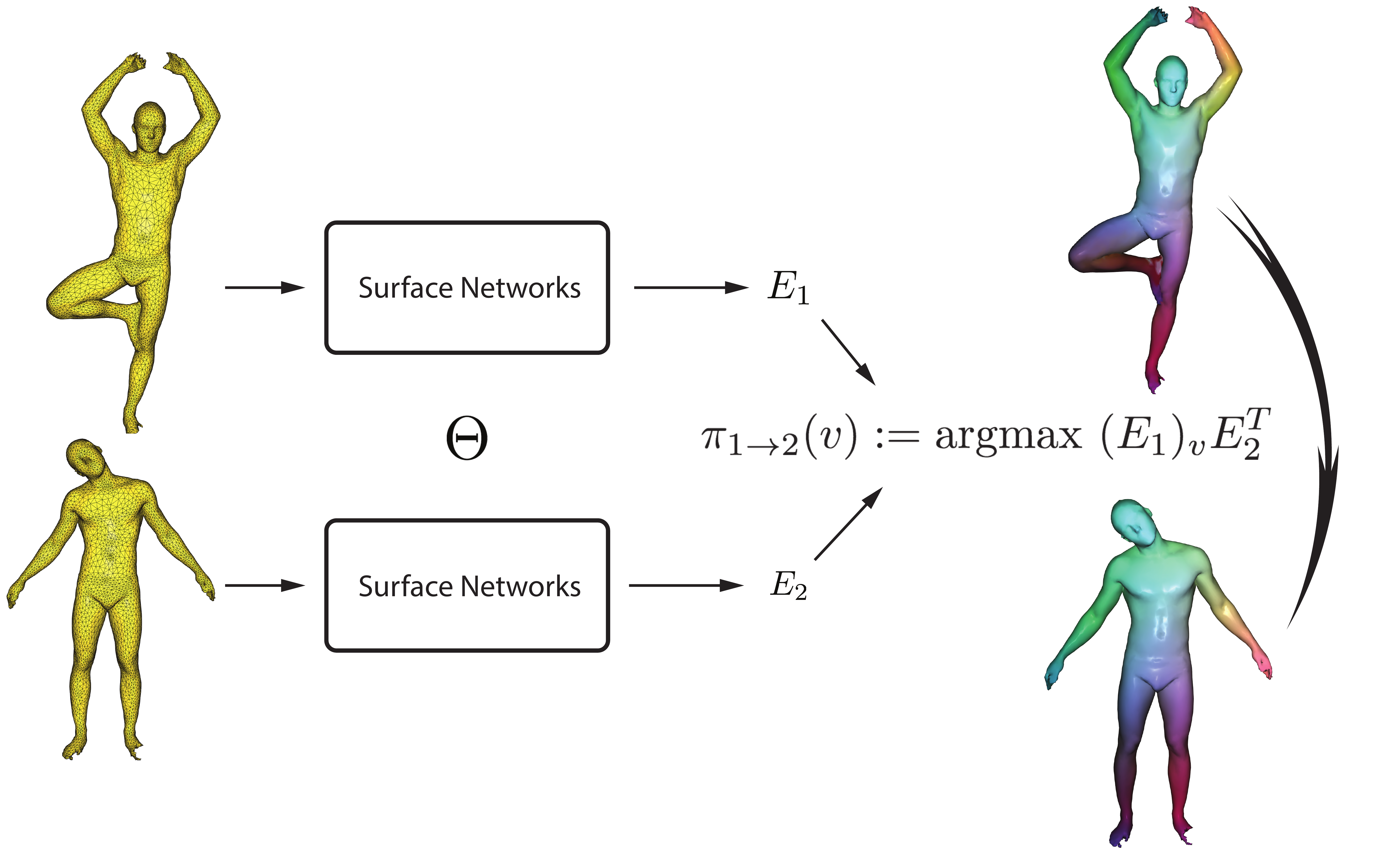}
    \caption{Siamese network pipeline: the two networks take vertex coordinates of the input models and generate a high dimensional feature vector, which are then used to define a map from $\mathcal{M}_1$ to $\mathcal{M}_2$. Here, the map is visualized by taking a color map on $\mathcal{M}_2$, and transferring it on $\mathcal{M}_1$ }
    \label{shapecorrfig1}
\end{figure}

Similarly as the graph correspondence model from \cite{nowak2017note}, 
we consider a Siamese Surface Network, 
consisting of two identical models with the same architecture and sharing parameters. 
For a pair of input surfaces $\mathcal{M}_1, \mathcal{M}_2$ of $N_1$, $N_2$ points respectively, 
the network produces embeddings $E_1 \in \R^{N_1 \times d}$ and $E_2 \in \R^{N_2 \times d}$. 
These embeddings define a trainable similarity between points given by
\begin{equation}
\label{formulaa}
s_{i,j} = \frac{e^{\langle E_{1,i}, E_{2,j} \rangle}}{\sum_{j'} e^{\langle E_{1,i}, E_{2,j'} \rangle}},    
\end{equation}

which can be trained by minimizing the cross-entropy relative to ground truth pairs. A diagram of the architecture is provided in Figure \ref{shapecorrfig1}.

In general, dense shape correspondence is a task that requires a blend of intrinsic and extrinsic information, motivating the use of data-driven models that can obtain such tradeoffs automatically. 
Following the setup in Experiment 6.2, we use models with 15 ResNet-v2 blocks with 128 output features each, and alternate Laplace and Dirac based models with Average Pooling blocks to cover a larger context: The input to our network consists of vertex positions only.

We tested our architecture on a reconstructed (i.e. changing the mesh connectivity) version of the real scan of FAUST dataset\cite{bogo2014faust}. The FAUST dataset contains 100 real scans and their corresponding ground truth registrations. The ground truth is based on a deformable template mesh with the same ordering and connectivity, which is fitted to the scans. In order to eliminate the bias of using the same template connectivity, 
as well as the need of a single connected component, 
the scans are reconstructed again with \cite{tetwild}. To foster replicability, we release the processed dataset in the additional material. In our experiment, we use 80 models for training and 20 models for testing.

Since the ground truth correspondence is implied only through the common template mesh, we compute the correspondence between our meshes with a  nearest neighbor search between the point cloud and the reconstructed mesh.
Consequently, due to the drastic change in vertex replacement after the remeshing, only 60-70 percent of labeled matches are used.
Although making it more challenging, we believe this setup is close to a real case scenario, where acquisition noise and occlusions are unavoidable.


\begin{figure}[h!]
\centering
    \includegraphics[width=\textwidth]{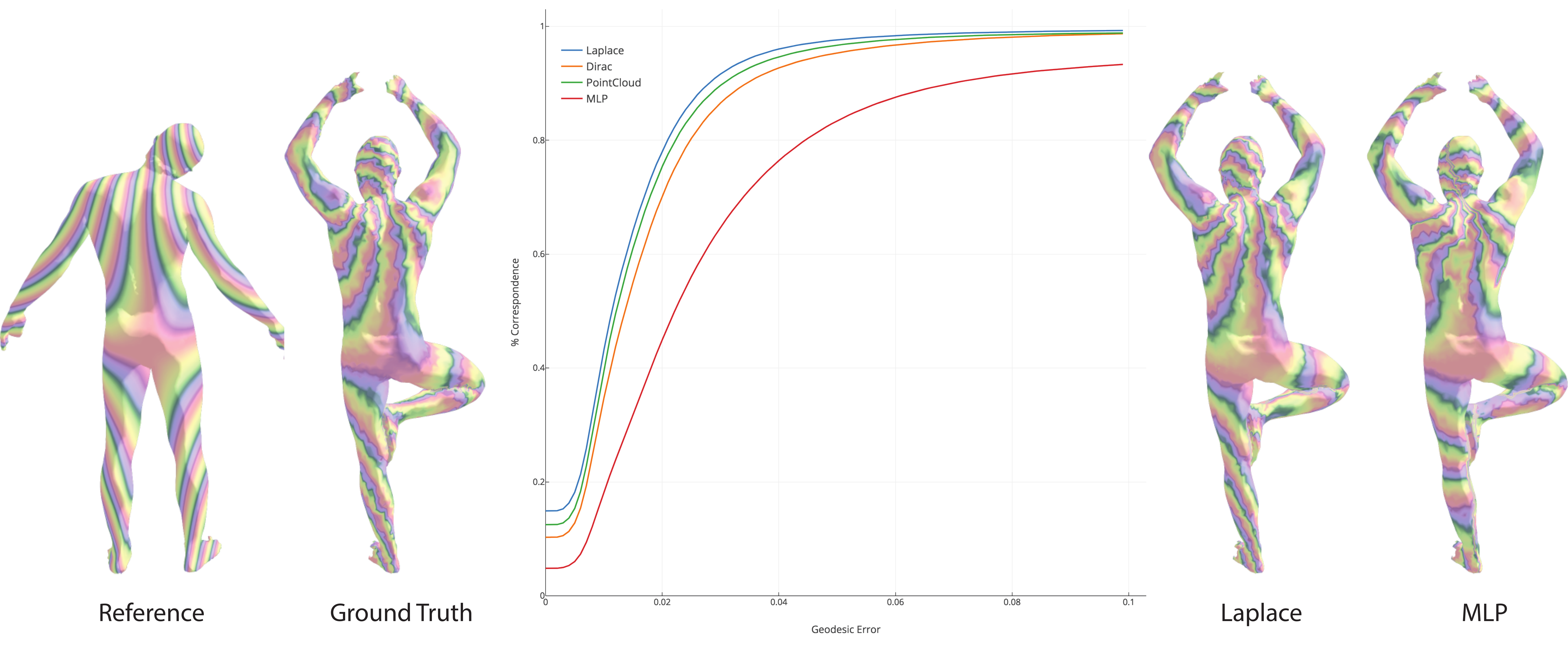} 
\caption{Additional results from our setup. Plot in the middle shows rate of correct correspondence with respect to geodesic error \cite{kim2011blended}. We observe that Laplace is performing similarly to Dirac in this scenario. We believe that the reason is that the FAUST dataset contains only isometric deformations, and thus the two operators have access to the same information.
We also provide visual comparison, with the transfer of a higher frequency colormap from the reference shape to another pose.}
\label{shapecorrfig2}
\end{figure}

\begin{figure}[h]
\centering
    \includegraphics[width=0.8\textwidth]{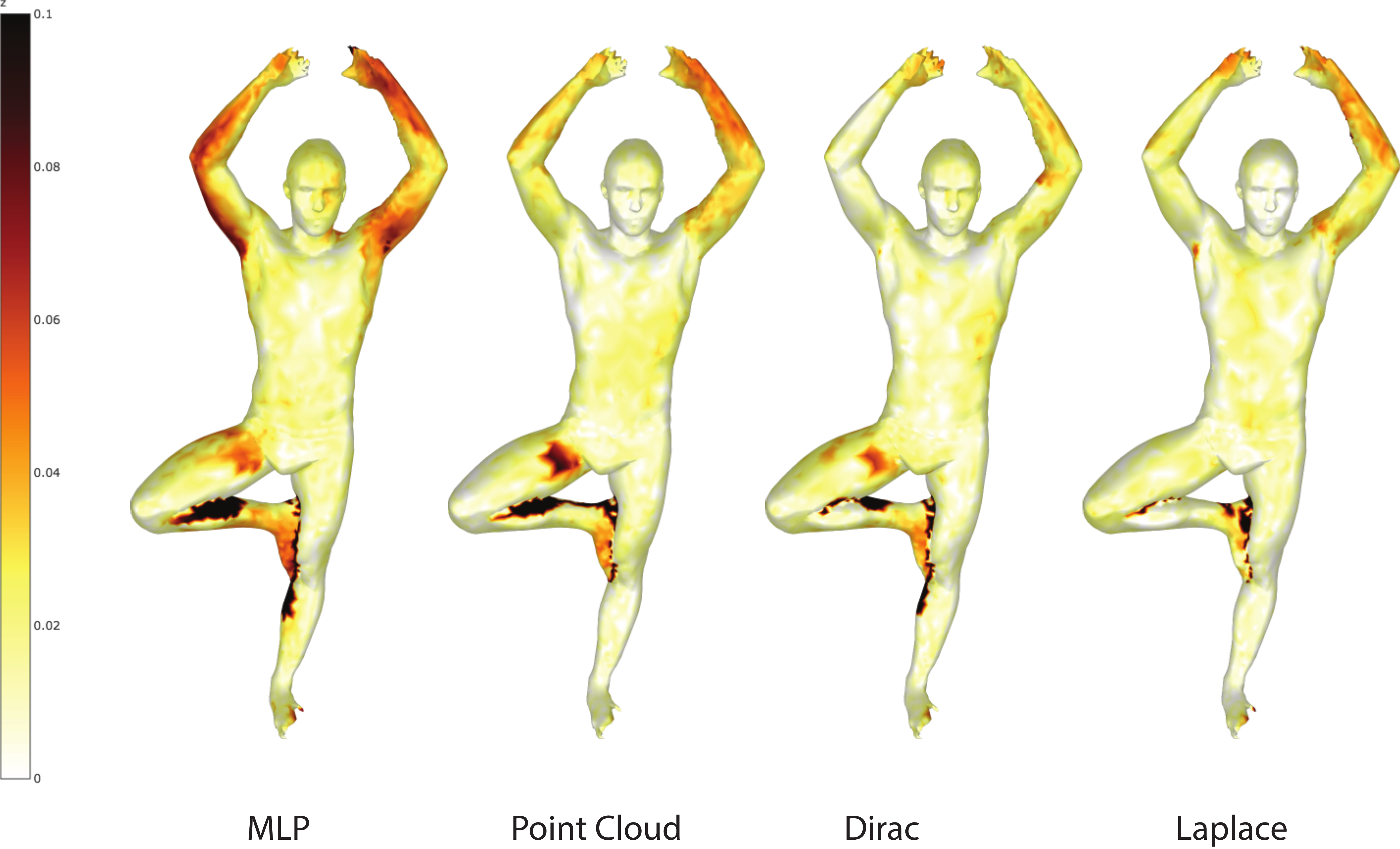} 
\caption{Heat map illustrating the point-wise geodesic difference between predicted correspondence point and the ground truth.
The unit is proportional to the geodesic diameter, and saturated at 10\%.}
\label{shapecorrfig_heatmap}
\end{figure}

\begin{figure}[h!]
\centering
    \includegraphics[width=0.8\textwidth]{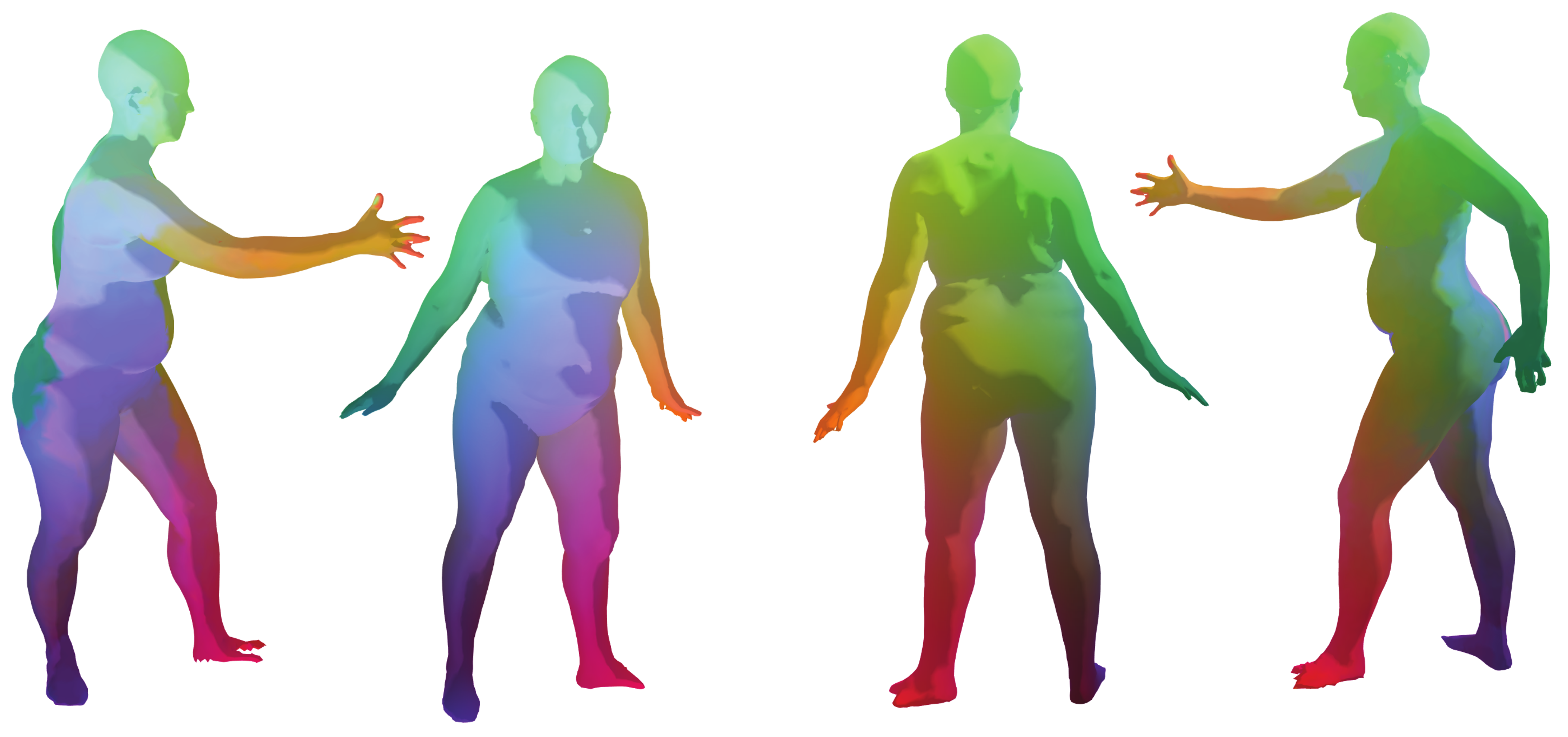} 
\caption{A failure case of applying the Laplace network to a new pose in the FAUST benchmark dataset. The network confuses between left and right arms. We show the correspondence visualization for front and back of this pair.}
\label{shapecorrfig3}
\end{figure}
Our preliminary results are reported in Figure \ref{shapecorrfig2}. For simplicity, we generate predicted correspondences by simply taking the mode of the softmax distribution for each reference node $i$: $\hat{j}(i) = \arg\max_j s_{i,j}$, thus avoiding a refinement step that is standard in other shape correspondence pipelines. 
The MLP model uses no context whatsoever and provides a baseline that captures the prior information from input coordinates alone. Using contextual information (even extrinsically as in point-cloud model) brings significative improvments, but these results may be substantially improved by encoding further prior knowledge. An example of the current failure of our model is depitcted in Figure \ref{shapecorrfig3}, illustrating that our current architecture does not have sufficiently large spatial context to disambiguate between locally similar (but globally inconsistent) parts. 

We postulate  that the FAUST dataset \cite{bogo2014faust} is not an ideal fit for our contribution for two reasons: (1) it is small (100 models), and (2) it contains only near-isometric deformations, which do not require the generality offered by our network. As demonstrated in \cite{Litany2017DeepFM}, the correspondence performances can be dramatically improved by constructing basis that are invariant to the deformations. 
We look forward to the emergence of new geometric datasets, and we are currently developing a capture setup that will allow us to acquire a more challenging dataset for this task.

  \section{Further Numerical Experiments}
\label{more_results}

\begin{figure}[ht!]
\centering
 \begin{tabular}{c c c c c} 
Ground Truth & MLP & AvgPool & Laplace & Dirac \\
 \includegraphics[width=0.19\textwidth]{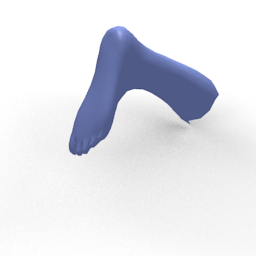} &
 \includegraphics[width=0.19\textwidth]{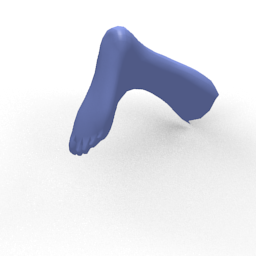} &
 \includegraphics[width=0.19\textwidth]{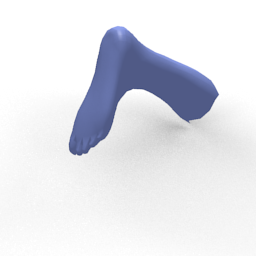} &
 \includegraphics[width=0.19\textwidth]{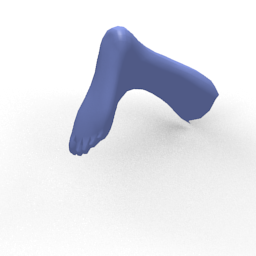} &
 \includegraphics[width=0.19\textwidth]{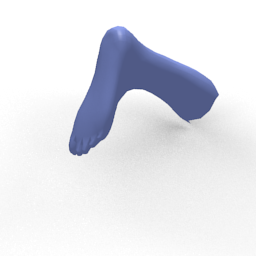} \\
 \includegraphics[width=0.19\textwidth]{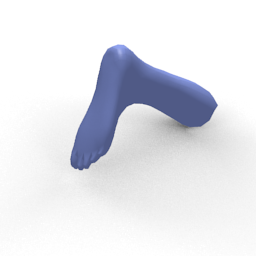} &
 \includegraphics[width=0.19\textwidth]{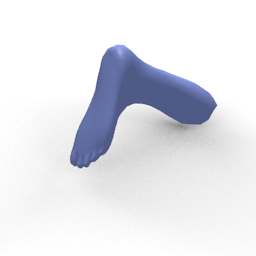} &
 \includegraphics[width=0.19\textwidth]{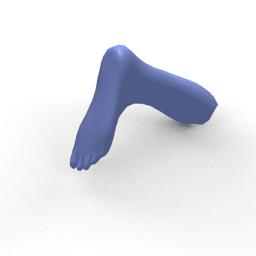} &
 \includegraphics[width=0.19\textwidth]{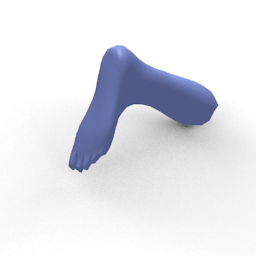} &
 \includegraphics[width=0.19\textwidth]{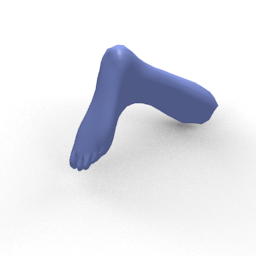} \\
 \includegraphics[width=0.19\textwidth]{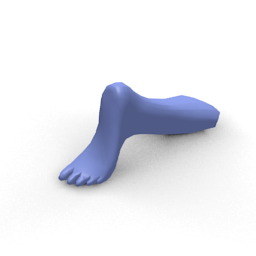} &
 \includegraphics[width=0.19\textwidth]{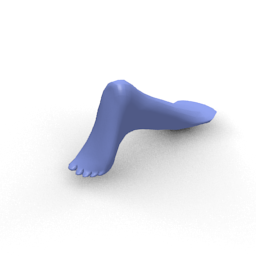} &
 \includegraphics[width=0.19\textwidth]{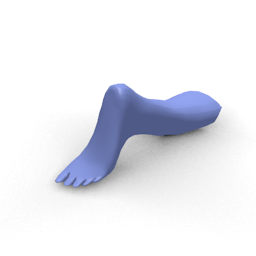} &
 \includegraphics[width=0.19\textwidth]{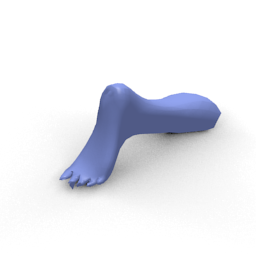} &
 \includegraphics[width=0.19\textwidth]{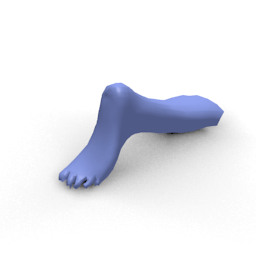} \\
 \includegraphics[width=0.19\textwidth]{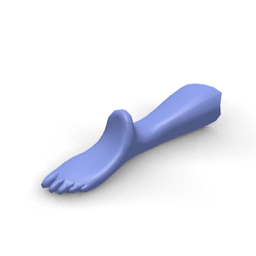} &
 \includegraphics[width=0.19\textwidth]{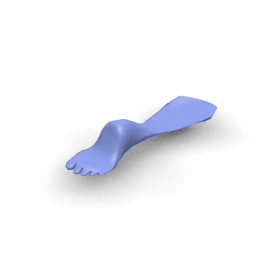} &
 \includegraphics[width=0.19\textwidth]{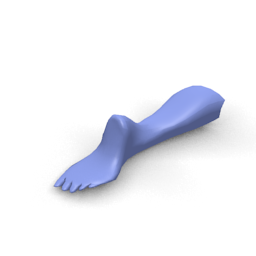} &
 \includegraphics[width=0.19\textwidth]{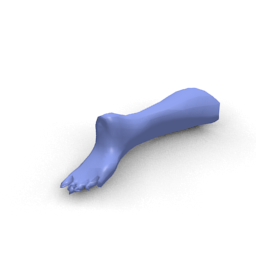} &
 \includegraphics[width=0.19\textwidth]{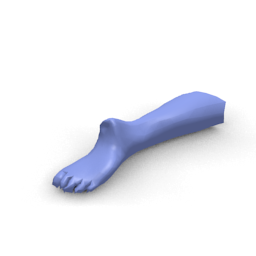} \\
 \includegraphics[width=0.19\textwidth]{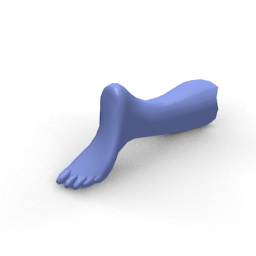} &
 \includegraphics[width=0.19\textwidth]{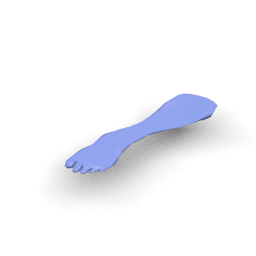} &
 \includegraphics[width=0.19\textwidth]{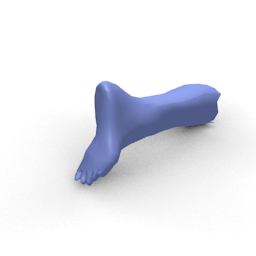} &
 \includegraphics[width=0.19\textwidth]{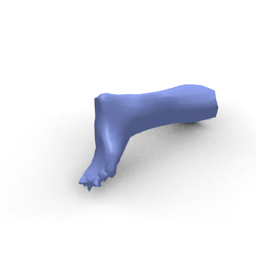} &
 \includegraphics[width=0.19\textwidth]{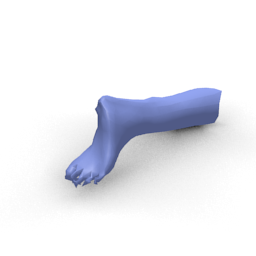} \\
 \end{tabular}
\caption{Qualitative comparison of different models. We plot 1th, 10th, 20th, 30th and 40th predicted frame correspondingly.}  
\end{figure}

\begin{figure}[ht!]
\centering
 \begin{tabular}{c c c c c} 
Ground Truth & MLP & AvgPool & Laplace & Dirac \\
 \includegraphics[width=0.19\textwidth]{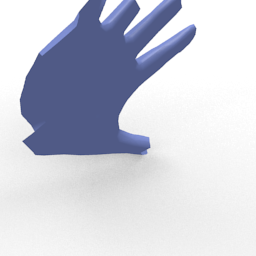} &
 \includegraphics[width=0.19\textwidth]{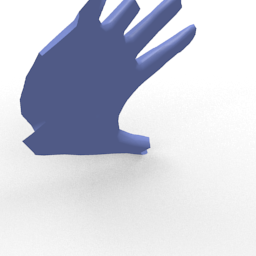} &
 \includegraphics[width=0.19\textwidth]{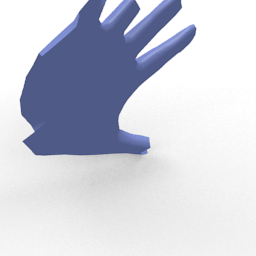} &
 \includegraphics[width=0.19\textwidth]{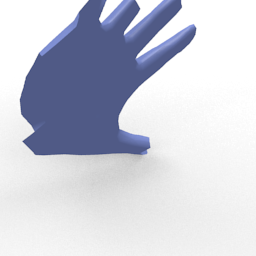} &
 \includegraphics[width=0.19\textwidth]{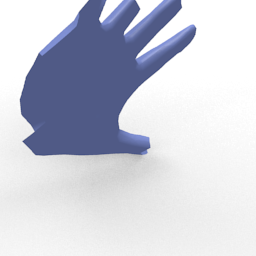} \\
 \includegraphics[width=0.19\textwidth]{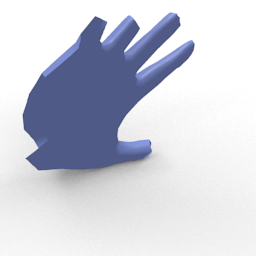} &
 \includegraphics[width=0.19\textwidth]{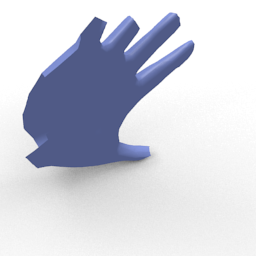} &
 \includegraphics[width=0.19\textwidth]{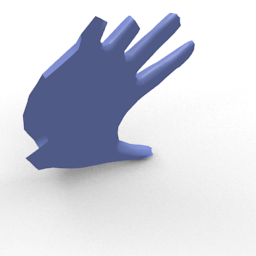} &
 \includegraphics[width=0.19\textwidth]{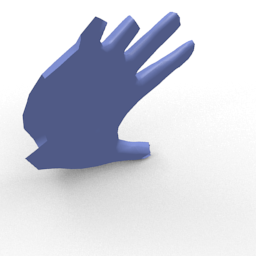} &
 \includegraphics[width=0.19\textwidth]{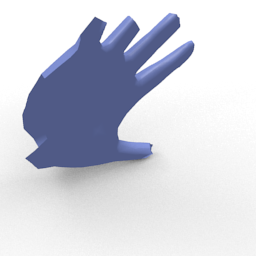} \\
 \includegraphics[width=0.19\textwidth]{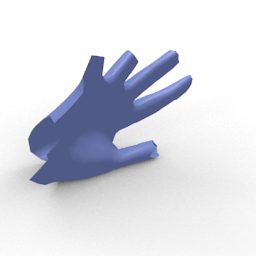} &
 \includegraphics[width=0.19\textwidth]{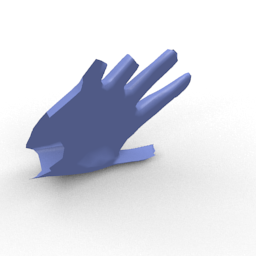} &
 \includegraphics[width=0.19\textwidth]{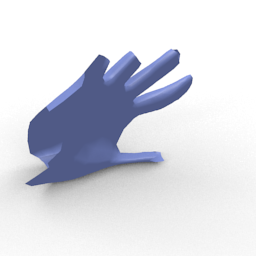} &
 \includegraphics[width=0.19\textwidth]{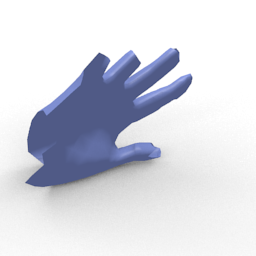} &
 \includegraphics[width=0.19\textwidth]{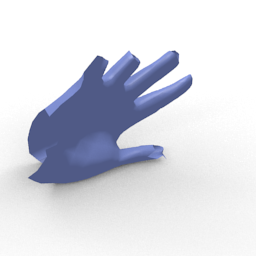} \\
 \includegraphics[width=0.19\textwidth]{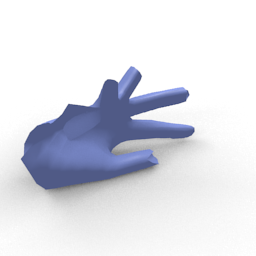} &
 \includegraphics[width=0.19\textwidth]{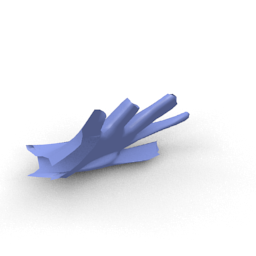} &
 \includegraphics[width=0.19\textwidth]{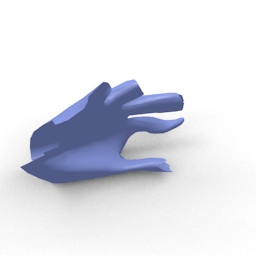} &
 \includegraphics[width=0.19\textwidth]{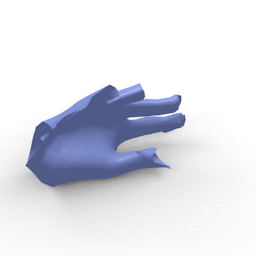} &
 \includegraphics[width=0.19\textwidth]{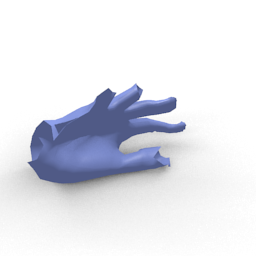} \\
 \includegraphics[width=0.19\textwidth]{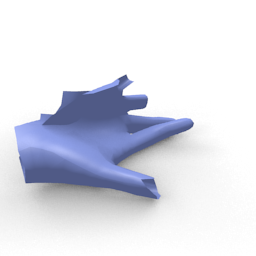} &
 \includegraphics[width=0.19\textwidth]{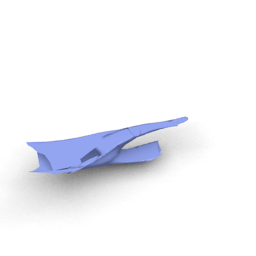} &
 \includegraphics[width=0.19\textwidth]{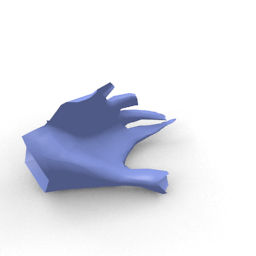} &
 \includegraphics[width=0.19\textwidth]{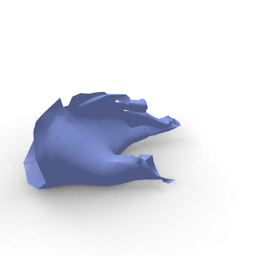} &
 \includegraphics[width=0.19\textwidth]{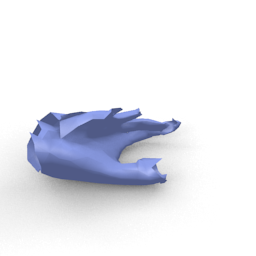} \\
 \end{tabular}
\caption{Qualitative comparison of different models. We plot 1th, 10th, 20th, 30th and 40th predicted frame correspondingly.}  
\end{figure}

\begin{figure}[ht!]
\centering
 \begin{tabular}{c c c c c} 
Ground Truth & MLP & AvgPool & Laplace & Dirac \\
 \includegraphics[width=0.19\textwidth]{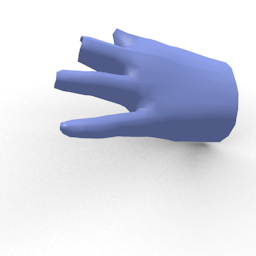} &
 \includegraphics[width=0.19\textwidth]{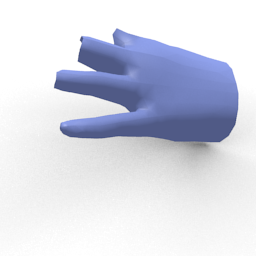} &
 \includegraphics[width=0.19\textwidth]{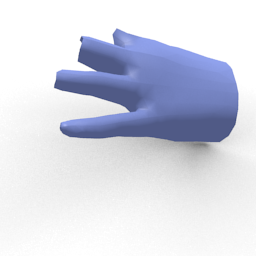} &
 \includegraphics[width=0.19\textwidth]{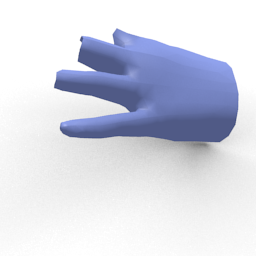} &
 \includegraphics[width=0.19\textwidth]{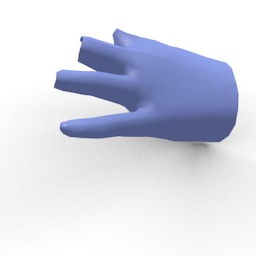} \\
 \includegraphics[width=0.19\textwidth]{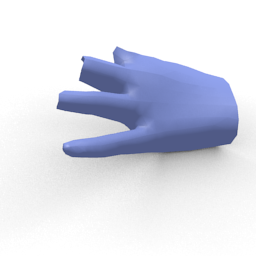} &
 \includegraphics[width=0.19\textwidth]{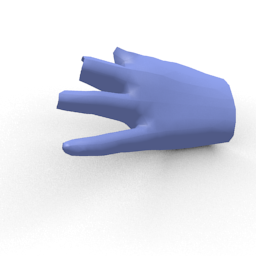} &
 \includegraphics[width=0.19\textwidth]{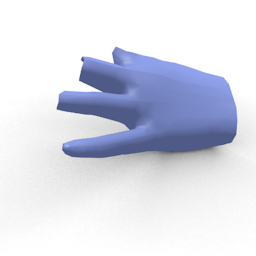} &
 \includegraphics[width=0.19\textwidth]{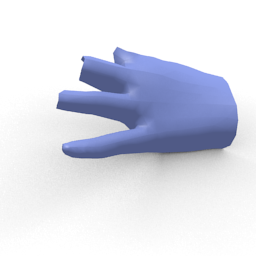} &
 \includegraphics[width=0.19\textwidth]{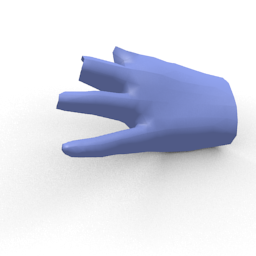} \\
 \includegraphics[width=0.19\textwidth]{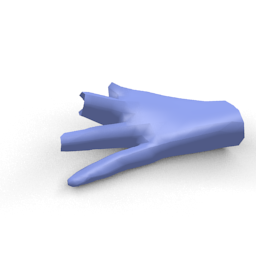} &
 \includegraphics[width=0.19\textwidth]{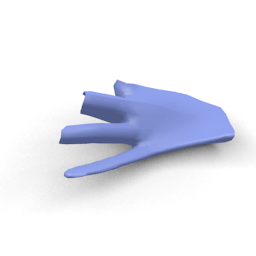} &
 \includegraphics[width=0.19\textwidth]{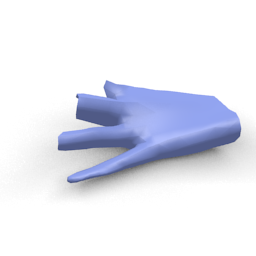} &
 \includegraphics[width=0.19\textwidth]{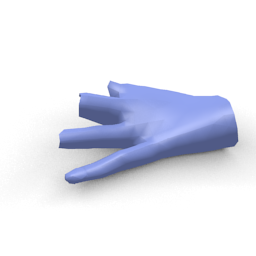} &
 \includegraphics[width=0.19\textwidth]{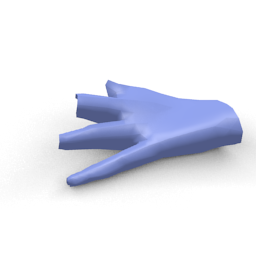} \\
 \includegraphics[width=0.19\textwidth]{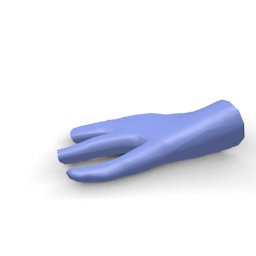} &
 \includegraphics[width=0.19\textwidth]{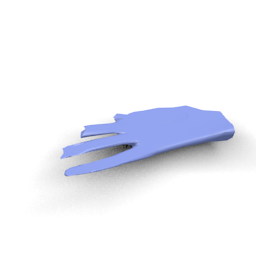} &
 \includegraphics[width=0.19\textwidth]{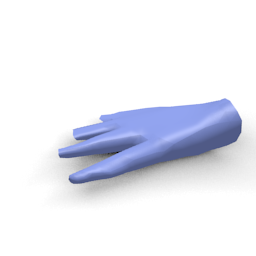} &
 \includegraphics[width=0.19\textwidth]{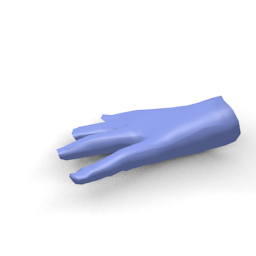} &
 \includegraphics[width=0.19\textwidth]{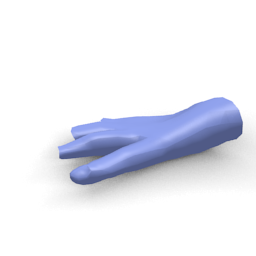} \\
 \includegraphics[width=0.19\textwidth]{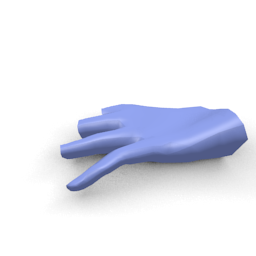} &
 \includegraphics[width=0.19\textwidth]{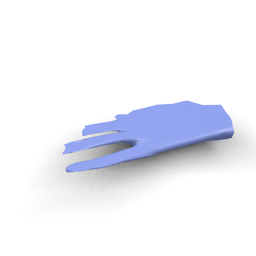} &
 \includegraphics[width=0.19\textwidth]{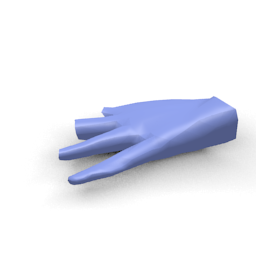} &
 \includegraphics[width=0.19\textwidth]{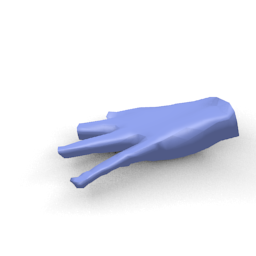} &
 \includegraphics[width=0.19\textwidth]{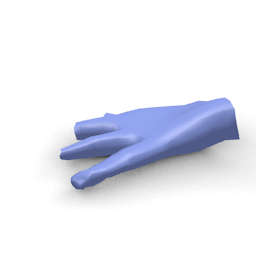} \\
 \end{tabular}
\caption{Qualitative comparison of different models. We plot 1th, 10th, 20th, 30th and 40th predicted frame correspondingly.}  
\end{figure}

\begin{figure}[ht!]
\centering
 \begin{tabular}{c c c c c} 
Ground Truth & MLP & AvgPool & Laplace & Dirac \\
 \includegraphics[width=0.19\textwidth]{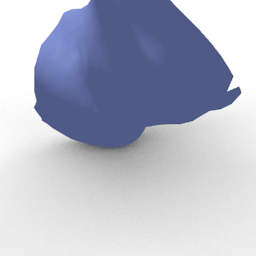} &
 \includegraphics[width=0.19\textwidth]{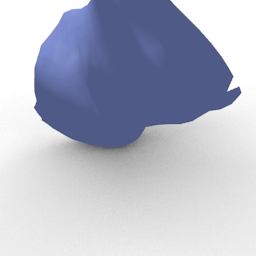} &
 \includegraphics[width=0.19\textwidth]{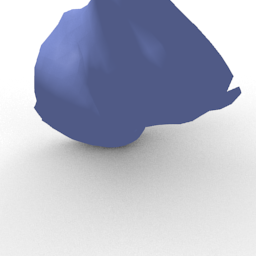} &
 \includegraphics[width=0.19\textwidth]{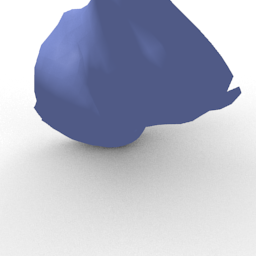} &
 \includegraphics[width=0.19\textwidth]{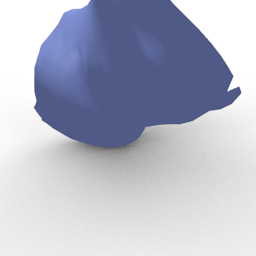} \\
 \includegraphics[width=0.19\textwidth]{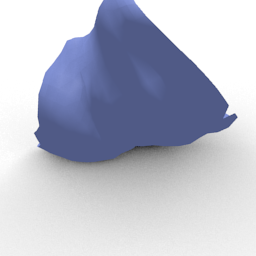} &
 \includegraphics[width=0.19\textwidth]{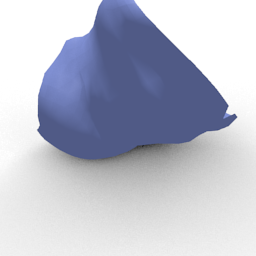} &
 \includegraphics[width=0.19\textwidth]{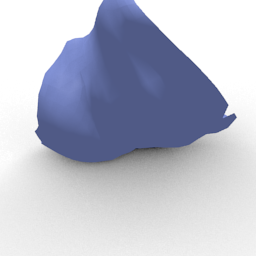} &
 \includegraphics[width=0.19\textwidth]{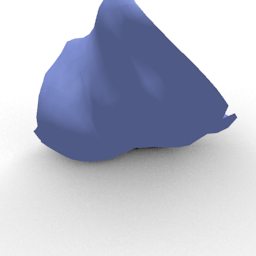} &
 \includegraphics[width=0.19\textwidth]{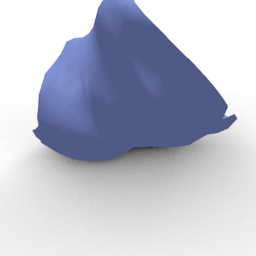} \\
 \includegraphics[width=0.19\textwidth]{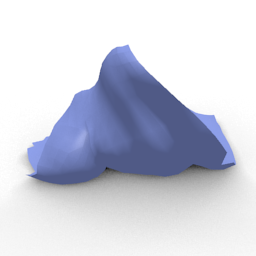} &
 \includegraphics[width=0.19\textwidth]{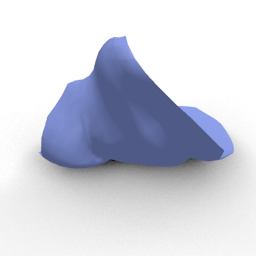} &
 \includegraphics[width=0.19\textwidth]{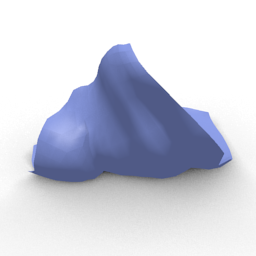} &
 \includegraphics[width=0.19\textwidth]{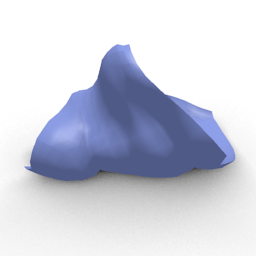} &
 \includegraphics[width=0.19\textwidth]{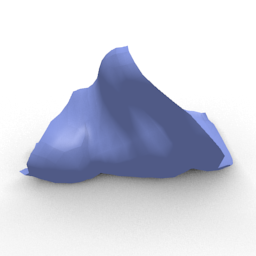} \\
 \includegraphics[width=0.19\textwidth]{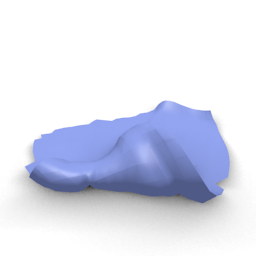} &
 \includegraphics[width=0.19\textwidth]{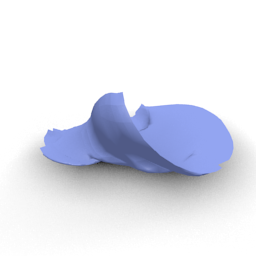} &
 \includegraphics[width=0.19\textwidth]{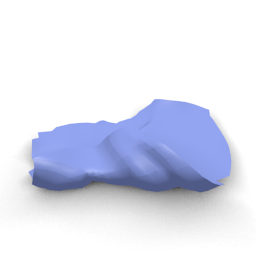} &
 \includegraphics[width=0.19\textwidth]{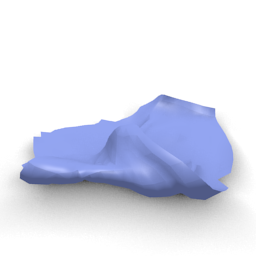} &
 \includegraphics[width=0.19\textwidth]{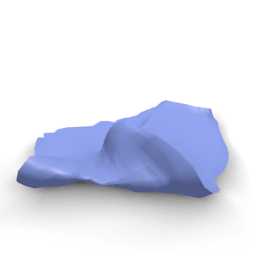} \\
 \includegraphics[width=0.19\textwidth]{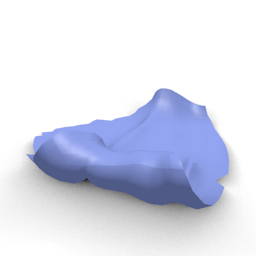} &
 \includegraphics[width=0.19\textwidth]{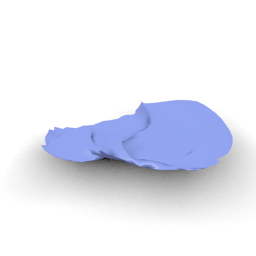} &
 \includegraphics[width=0.19\textwidth]{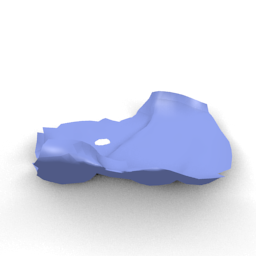} &
 \includegraphics[width=0.19\textwidth]{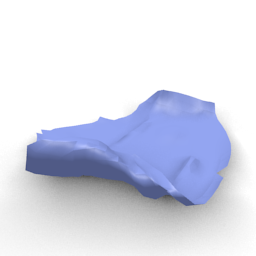} &
 \includegraphics[width=0.19\textwidth]{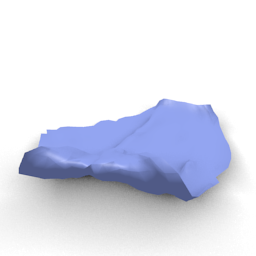} \\
 \end{tabular}
\caption{Qualitative comparison of different models. We plot 1th, 10th, 20th, 30th and 40th predicted frame correspondingly.}  
\end{figure}

\begin{figure}[ht!]
\centering
 \begin{tabular}{c c c} 
 Ground Truth & Laplace & Dirac \\
 \includegraphics[width=0.30\textwidth]{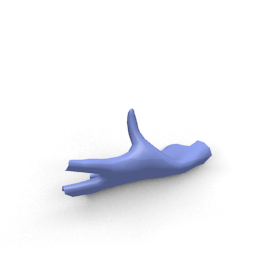} &
 \includegraphics[width=0.30\textwidth]{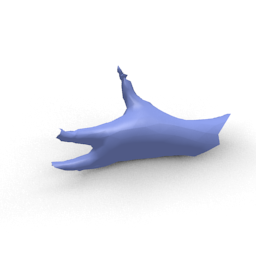} &
 \includegraphics[width=0.30\textwidth]{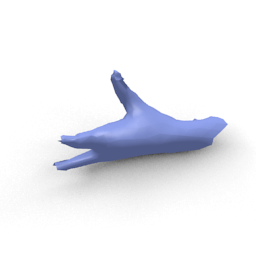} \\
 \includegraphics[width=0.30\textwidth]{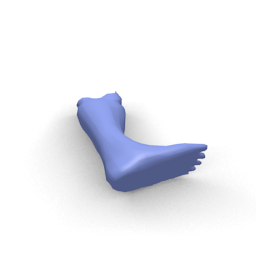} &
 \includegraphics[width=0.30\textwidth]{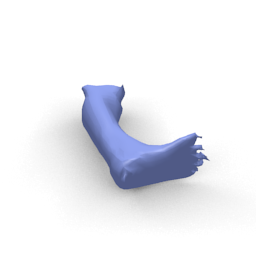} &
 \includegraphics[width=0.30\textwidth]{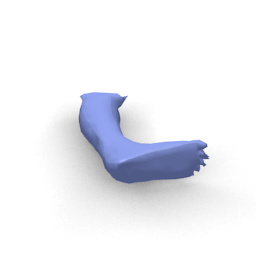}
 \end{tabular}
\caption{Dirac-based model visually outperforms Laplace-based models in the regions of high mean curvature.}  
\end{figure}

\begin{figure}[ht!]
\centering
\includegraphics[width=0.85\textwidth]{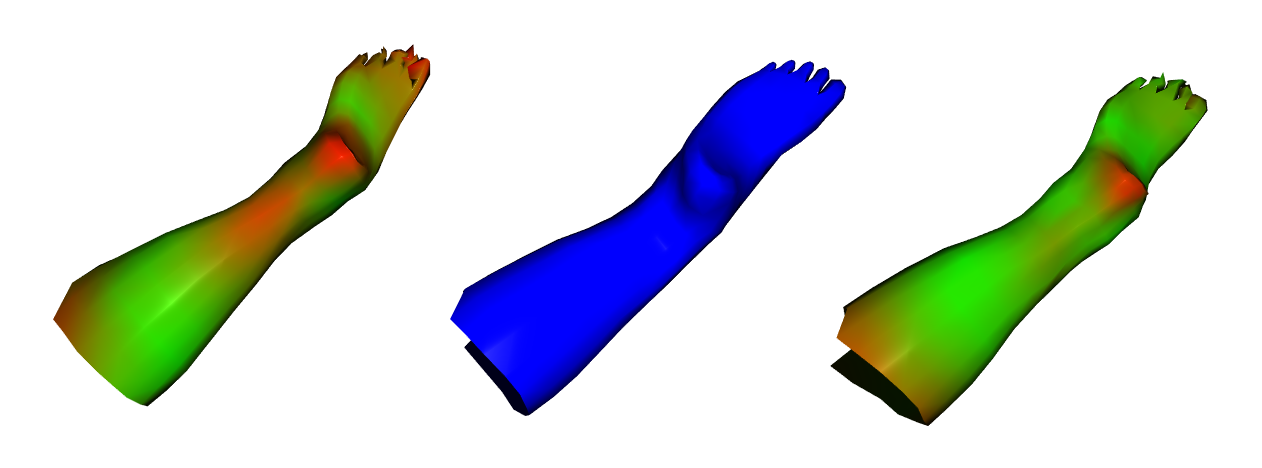} \\
\includegraphics[width=0.85\textwidth]{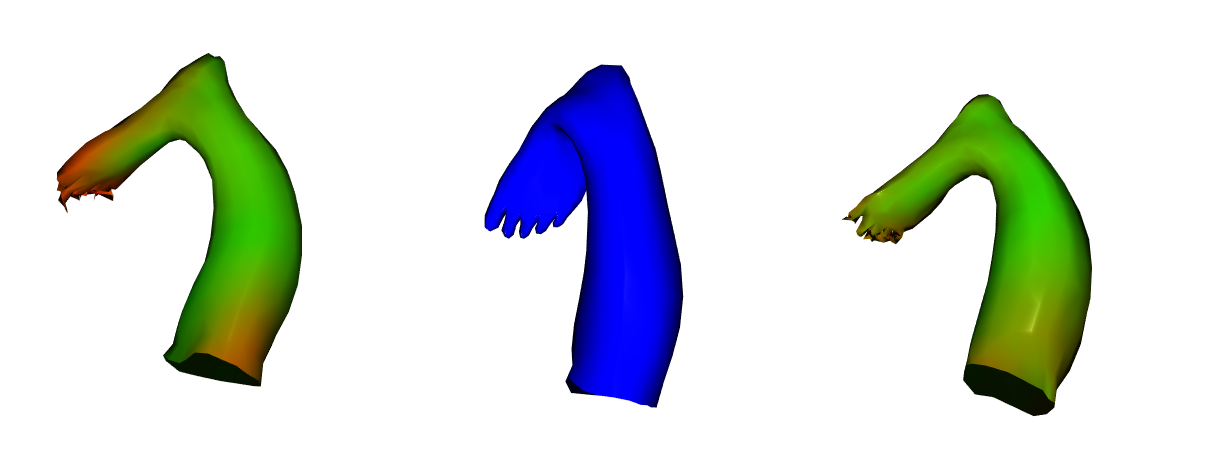} \\
\includegraphics[width=0.85\textwidth]{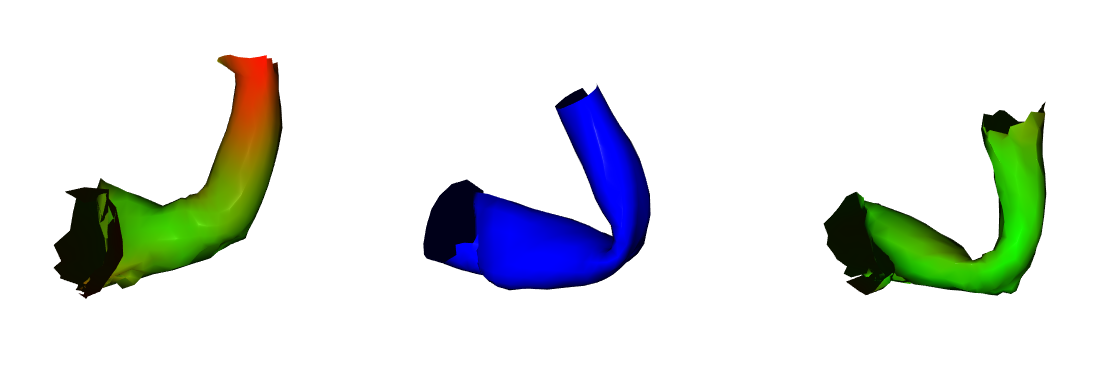}
\includegraphics[width=0.04\textwidth]{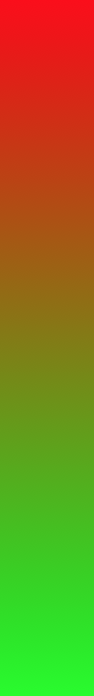}
\caption{From left to right: Laplace, ground truth and Dirac based model. Color corresponds to mean squared error between ground truth and prediction: green - smaller error, red - larger error.}  
\end{figure}

\begin{figure}[ht!]
\centering
\includegraphics[width=0.85\textwidth]{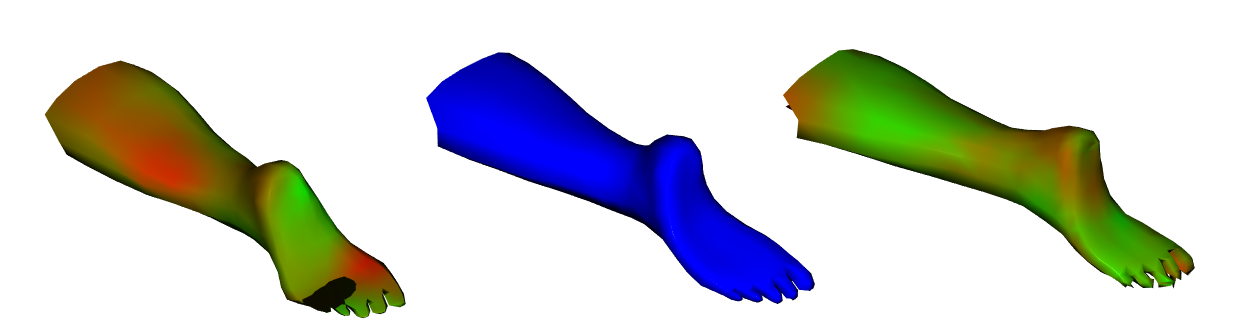} \\
\includegraphics[width=0.85\textwidth]{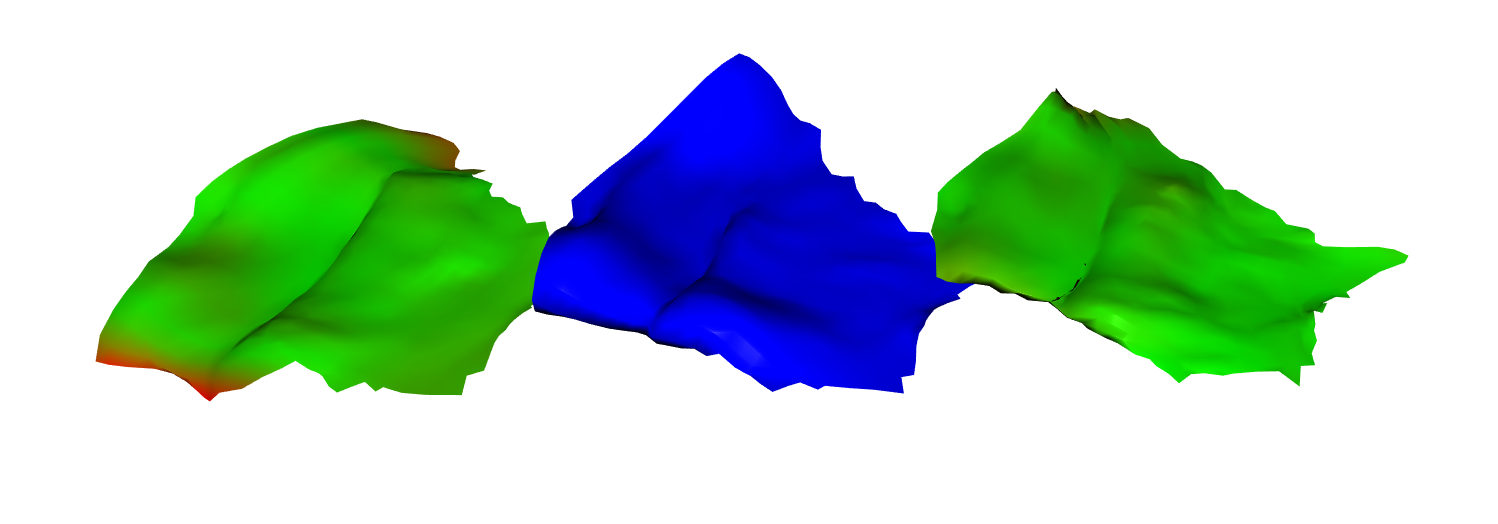} \\
\includegraphics[width=0.85\textwidth]{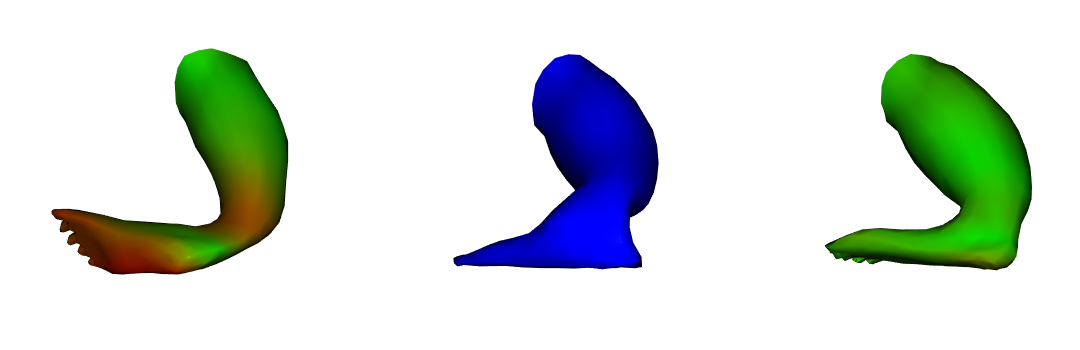}
\includegraphics[width=0.04\textwidth]{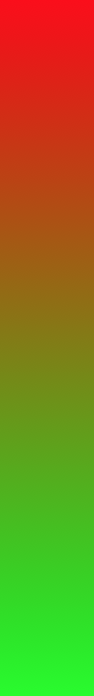}
\caption{From left to right: set-to-set, ground truth and Dirac based model. Color corresponds to mean squared error between ground truth and prediction: green - smaller error, red - larger error.}  
\end{figure}

\end{document}